\documentclass[journal]{elsarticle}

\usepackage{scalerel}
\usepackage{tikz}
\usetikzlibrary{svg.path}
\definecolor{orcidlogocol}{HTML}{A6CE39}
\tikzset{
    orcidlogo/.pic={
        \fill[orcidlogocol] svg{M256,128c0,70.7-57.3,128-128,128C57.3,256,0,198.7,0,128C0,57.3,57.3,0,128,0C198.7,0,256,57.3,256,128z};
        \fill[white] svg{M86.3,186.2H70.9V79.1h15.4v48.4V186.2z}
        svg{M108.9,79.1h41.6c39.6,0,57,28.3,57,53.6c0,27.5-21.5,53.6-56.8,53.6h-41.8V79.1z M124.3,172.4h24.5c34.9,0,42.9-26.5,42.9-39.7c0-21.5-13.7-39.7-43.7-39.7h-23.7V172.4z}
        svg{M88.7,56.8c0,5.5-4.5,10.1-10.1,10.1c-5.6,0-10.1-4.6-10.1-10.1c0-5.6,4.5-10.1,10.1-10.1C84.2,46.7,88.7,51.3,88.7,56.8z};
    }
}

\newcommand\orcidicon[1]{\href{https://orcid.org/#1}{\mbox{\scalerel*{
                \begin{tikzpicture}[yscale=-1,transform shape]
                \pic{orcidlogo};
                \end{tikzpicture}
            }{|}}}}

\usepackage{amssymb}
\usepackage{graphicx}
\usepackage{amsmath}
\usepackage{adjustbox}
\usepackage{multirow}
\usepackage{mathtools}	
\usepackage{varwidth}
\usepackage{tabularx}
\usepackage{siunitx}
\usepackage{url}
\usepackage{mathtools}
\usepackage{commath}
\usepackage{pdflscape}
\usepackage{siunitx}
\usepackage{enumitem}
\usepackage{pdflscape}
\usepackage[linesnumbered,ruled,vlined]{algorithm2e}
\usepackage{bm}
\usepackage{subcaption}

\usepackage{booktabs}

\usepackage{makecell}

\newcolumntype{R}[2]{%
	>{\adjustbox{angle=#1,lap=\width-(#2)}\bgroup}%
	l%
	<{\egroup}%
}
\usepackage[utf8]{inputenc}
\usepackage[T1]{fontenc}
\usepackage[english]{babel}
\usepackage{lmodern}

\usepackage{array}
\newcommand{\PreserveBackslash}[1]{\let\temp=\\#1\let\\=\temp}
\newcolumntype{C}[1]{>{\PreserveBackslash\centering}p{#1}}
\newcolumntype{R}[1]{>{\PreserveBackslash\raggedleft}p{#1}}
\newcolumntype{L}[1]{>{\PreserveBackslash\raggedright}p{#1}}

\journal{Neural Networks}
\usepackage{xcolor,colortbl}
\definecolor{migris}{rgb}{0.78,0.78,0.78}
\newcommand{\mejor}{\cellcolor{migris}}  %{0.9}
\definecolor{migris_2}{rgb}{0.92,0.92,0.92}
\newcommand{\seg}{\cellcolor{migris_2}}  %{0.9}
\begin{document}

\begin{frontmatter}

%\title{Detection and Attribution of Out-of-Distribution Samples in Spiking Neural Networks}
\title{A Novel Explainable Out-of-Distribution Detection Approach for Spiking Neural Networks}

\author[address1,address2]{Aitor Martinez Seras\corref{mycorrespondingauthor}}
\cortext[mycorrespondingauthor]{Corresponding author. Parque Tecnologico de Bizkaia, 700, 48160 Derio, Bizkaia, Spain. Telephone: +34946430850}
\ead{aitor.martinez@tecnalia.com}

\author[address1,address3]{Javier~Del~Ser}

\author[address1]{Jesus L. Lobo}
\author[address2]{\\Pablo Garcia-Bringas}
\author[address4,address5,address6]{Nikola Kasabov}

\address[address1]{TECNALIA, Basque Research and Technology Alliance (BRTA), 48160 Derio, Spain}
\address[address2]{University of Deusto, 48007 Bilbao, Spain}
\address[address3]{University of the Basque Country (UPV/EHU), 48013 Bilbao, Spain}
\address[address4]{Auckland University of Technology, Auckland, New Zealand}
\address[address5]{Intelligent Systems Research Center, Ulster University, UK}
\address[address6]{IICT, Bulgarian Academy of Sciences, Bulgaria\vspace{-6mm}}

\begin{abstract}
Research around Spiking Neural Networks has ignited during the last years due to their advantages when compared to traditional neural networks, including their efficient processing and inherent ability to model complex temporal dynamics. Despite these differences, Spiking Neural Networks face similar issues than other neural computation counterparts when deployed in real-world settings. This work addresses one of the practical circumstances that can hinder the trustworthiness of this family of models: the possibility of querying a trained model with samples far from the distribution of its training data (also referred to as Out-of-Distribution or OoD data). Specifically, this work presents a novel OoD detector that can identify whether test examples input to a Spiking Neural Network belong to the distribution of the data over which it was trained. For this purpose, we characterize the internal activations of the hidden layers of the network in the form of spike count patterns, which lay a basis for determining when the activations induced by a test instance is atypical. Furthermore, a local explanation method is devised to produce attribution maps revealing which parts of the input instance push most towards the detection of an example as an OoD sample. Experimental results are performed over several image classification datasets to compare the proposed detector to other OoD detection schemes from the literature. As the obtained results clearly show, the proposed detector performs competitively against such alternative schemes, and produces relevance attribution maps that conform to expectations for synthetically created OoD instances.
\end{abstract}


\end{frontmatter}

%\linenumbers

\section{Introduction}
\label{sec:Introduction}

In the last decades, Machine Learning (ML) has been widely proven to outperform human capabilities in tasks that were previously thought to be intractable for machines, such as image/video classification, natural language processing or generative modeling. The unprecedented levels of performance achieved over these tasks have spurred an ever-growing proliferation of practical applications exploiting the capabilities of ML models. The sensitive nature of practical decisions that can be made as per the output elicited by ML models has lately motivated a flurry of research around the need for ensuring the trustworthiness of the audience consuming their inputs. This issue is particularly concerning when it comes to models whose internal structure and/or learning algorithm cannot be easily understood by non-expert users. This is the case of Deep Learning, arguably the field where the most outstanding breakthroughs have been reported in manifold domains \cite{jumper2021highly, brown2020language}. Concerns with the trustworthiness of Deep Learning models are even more acute as a result of their inherent black-box nature. Consequently, a new field of study has developed over the last decade, coined as eXplainable Artificial Intelligence~(XAI), which focuses on making the decisions issued by ML models understandable for humans \cite{arrieta2020explainable, adadi2018peeking, tjoa2020survey}. 

Besides the lack of interpretability of ML models, other factors may hinder the trustworthiness of ML models when deployed in real-world contexts. In such scenarios, models can face more varying and complex circumstances than the ones encountered in the controlled environments where their parameters are usually learned. Among them, one of the essential assumptions made in ML is that both training and test data are drawn from the same distribution. However, this assumption may not hold in real-world settings, in which this distribution may change, and test instances might be sampled from a unknown distribution. In general ML models are not designed from scratch to discriminate between \emph{known} and \emph{unknown} query samples as per the knowledge captured during their training processes. In this case, it becomes necessary to endow the model with the ability to determine whether the queried sample belongs to the data distribution it has learned. This task is exactly what Out-of-Distribution (OoD) detection addresses \cite{yang2021generalized}: unlike outlier detection, OoD methods depart from a model already trained over a given dataset, namely, the \emph{In-Distribution} (ID) dataset. The goal pursued in OoD detection is to declare whether a new test instance has either been sampled from the modeled ID dataset or, instead, it belongs to another distribution (correspondingly, the \emph{Out-Distribution} dataset, OD). Given its straightforward utility in uncontrolled environments, OoD detection has been acknowledged to be of pivotal importance in many applications, including safe autonomous driving \cite{muhammad2020deep} and medical diagnosis \cite{uwimana2021out,cao2020benchmark}.

From a modeling perspective, the literature has tackled the OoD detection problem by focusing mainly on neural networks for image classification, using convolutional architectures for the purpose. In general, OoD instances can be identified by adopting i) model-agnostic detection methods; or ii) model-specific strategies. The former usually operate by characterizing information elicited by the model during its training process that is not particular of the model under target (e.g. the produced output class distribution of training samples). Model-specific methods, however, inspect the internal parameters of the model where its captured knowledge is persisted (e.g. forward-propagated activations of neurons). Above all, most detectors reported so far by the community remain as opaque as the models they aim to supervise, missing to attribute which parts of the input test instance are most influential for the detection of the whole sample as an OoD instance. In other words, OoD detection must be also explained for the audience of the model; otherwise, it could jeopardize the completion of a fully trustworthy ML pipeline. Another gap in the current literature is the lack of model-specific proposals suitable to detect OoD instances in Spiking Neural Networks (SNNs). These models, often referred to as \textit{the third generation of artificial neural networks}, are the evolution of present-day neural networks in virtue of their energy efficiency when implemented in specialized neuro-morphic hardware, and their ability to model complex temporal dynamics thanks to their event-based nature \cite{merolla2014million, ghosh2009third}. Despite the intense research activity around this family of models noted in recent times, no prior work exists dealing with the detection of OoD instances by leveraging specifics of the spike-based working procedure of SNNs.

This work covers the gap identified above in OoD research by developing a novel model-specific detection technique specifically designed for SNNs. Our proposed approach utilizes information produced by the SNN learning algorithm during the training process to build a representation of its typicality when processing ID data. A second contribution of our work is a local attribution method that highlights the parts of the input instance that push the developed detector towards classifying a sample as OoD. The attribution information produced by this method can improve the acceptability of the audience of the model when being notified about an \emph{unknown} input, connecting clearly to the trustworthiness of ML pipelines sought in risk-sensitive domains as the ones exemplified previously. An extensive experimental setup over several image classification tasks is designed i) to compare the devised OoD detection method to existing alternatives that can be adapted to SNNs; and ii) to assess whether the generated attributions succeed at identifying artificially induced image artifacts on the considered datasets. Results are promising, exposing a very competitive detection performance and attributions that conform to expectations.

The rest of the paper is structured as follows: first, Section \ref{sec:RelatedWork} contextualizes the OoD detection task, formally stating the problem it aims to solve, and introducing a taxonomy that allows framing the contribution of the work. This section also briefly mentions recent advances made in SNNs and XAI. Next, Section \ref{sec:ProposedApproach} presents the proposed detector by describing its overall workflow, followed by a detailed description of the detection algorithm and the procedure to extract local explanations in the form of attribution maps. The experimental setup is described in Section \ref{sec:Experiments}, whereas results are discussed quantitatively and qualitatively in Section \ref{sec:Results}. Finally, conclusions are drawn in Section \ref{sec:Conclusions}, together with several research paths rooted on our investigations.

\section{Related Work}
\label{sec:RelatedWork}

This study lies in the intersection between several fields of research. Consequently, this section briefly pauses at the state of the art of each of such fields: the OoD detection problem (Subsection \ref{ssec:OoD_related}), Spiking Neural Networks (Subsection \ref{ssec:SNNsRelated}) and Explainable AI in SNNs (Subsection \ref{ssec:xAI_in_SNNs}). Finally, we end by stating the contribution of this work beyond the current status of knowledge in these surveyed areas (Subsection \ref{ssec:contribution}).

\subsection{Out-of-distribution detection}
\label{ssec:OoD_related}

Before formulating the problem in a mathematical fashion, it is important to clearly define what the out-of-distribution detection encompasses and to establish the similarities and differences with other related tasks, such as outlier, anomaly and novelty detection. This is precisely the central matter tackled in the comprehensive survey recently contributed in~\cite{yang2021generalized}, which also points out several misconceptions existing in the literature. This work distinguishes between two different types of shift in data: the \emph{semantic shift} and the \emph{covariate shift}. The latter stands for the situation where features of the input data vary without any change in the class they belong to. Stated differently, the statistical dependence between variables change, but not their relationship with the output variable. In contrast, semantic shift refers to the case when the new data belongs to a class different than the ones learned by the model during its training stage. This being said, it is important to note that OoD detection addresses only semantic shift.
\begin{figure}[hbt]
    \centering
    \includegraphics[width=0.90\textwidth]{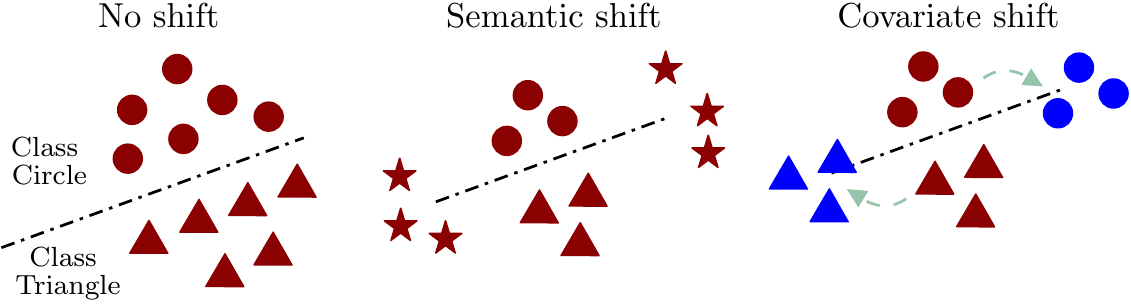}
    \caption{Difference between semantic and covariate shift. The semantic one occurs when a new class (star) emerges. In the covariate shift, the features of the input change (the color of the figures, from red to blue) while maintaining the semantics (the label), what makes the model incorrectly classify some instances.}
    \label{fig:SemanticVScovariate}
\end{figure}

To clarify the application domain in which OoD detection can be formulated, we start by differentiating it from the above-mentioned tasks. The most distinct task is outlier detection, as it does not follow the classic train-test schema. Instead, a set of observations is provided, and the task is to detect significantly different samples from this contaminated dataset of observations. Both semantic and covariate shifts can occur in this scenario. Conversely to outlier detection, OoD detection is assumed to occur under a train-test schema. As concluded in \cite{yang2021generalized}, OoD detection can be regarded as a super-category that includes semantic anomaly detection, multi-class novelty detection and open set recognition~\cite{geng2020recent}. Summarizing, in OoD detection a model is learned from a training set that is sampled from a given ID distribution, whereas test samples are drawn evenly from either the in-distribution or from a semantically different distribution, namely, the aforementioned OD dataset. The objective is to correctly classify test samples as belonging to the ID or OD distributions, while maintaining the performance of the model in the original classification task. In almost all cases, detection is ultimately achieved by crafting a scoring function that assigns different scores for the in- and out-distribution samples. 

Mathematically, the OoD detection task aims to distinguish whether a query instance $\mathbf{x}_{query}$ belongs to the data distribution of the data used to train the model, or alternatively can be thought to belong to another different distribution. Without loss of generality, let the \emph{in-distribution} dataset be denoted as $\mathcal{D}_{tr}$, which is drawn from a probability distribution $P_{\mathbf{X}}(\mathbf{x})$ defined over a space $\mathcal{X}\in\mathbb{R}^K$. This dataset is used to learn a classifier $M_{\theta}(\mathbf{x})$ by adjusting the model's parameters $\theta$ so as to maximize its generalization capability over a set of labels $\{1,\ldots,\ell\}$. Once the model has been learned, the classifier can infer the class $\hat{y}=M_\theta(\mathbf{x})\in\{1,\ldots,\ell\}$ of any new sample $\mathbf{x}\in\mathbb{R}^K$. We further assume another distribution $Q_{\mathbf{X}}(\mathbf{x})$ different than $P_{\mathbf{X}}(\mathbf{x})$, yet over the same space $\mathcal{X}$ (\emph{out-distribution}). New query samples can be sampled from a mixture distribution $M_{\mathbf{X}|Z}(\mathbf{x}|z)$ such that $M_{\mathbf{X}|Z}(\mathbf{x}|z)$ is equal to $P_{\mathbf{X}}(\mathbf{x})$ if $z=1$, and $Q_{\mathbf{X}}(\mathbf{x})$ if $z=0$. If $Z$ follows a uniform binary probability distribution, an OoD detector $G_{\varphi}(\mathbf{x})$ with parameters $\varphi$ aims to distinguish which test instances follow the in-distribution from the ones which not, without observing variable $Z$, and by only resorting to the trained model $M_\theta(\mathbf{x})$.

Recalling to \cite{yang2021generalized} and following its proposed taxonomy, OoD detection methods can be divided in three types: classification-based, density-based or distance-based. Our technique falls within the latter category, as it comprises a detector that relies on the computation of a measure of distance between the query samples and the centroids or prototypes of the classes present in the in-distribution. Nevertheless, we herein focus on the literature related to some of the classification-based methods, specifically those which can be labeled as \emph{post-hoc} methods. These detectors can be easily adapted to Spiking Neural Networks without making changes to their core, by solely resorting to information produced at their output (e.g., the logits).

The first paper coining the \emph{Out of Distribution detection} term was actually a classification-based method, presented in \cite{hendrycks2016baseline} and often referred to as the \emph{baseline} method. This approach utilizes the so-called Maximum Softmax Probability (MSP) to detect OoD samples, based on the observation that samples belonging to the ID dataset tend to have a higher MSP value than OD instances. By simply defining a threshold on this score, their method achieved acceptable results in a wide variety of datasets. The authors in \cite{liang2017odin} went one step further by applying a temperature scaling strategy to the softmax computation, which pushes softmax scores of in- and out-distribution samples further apart from each other. In addition, they included input preprocessing by adding small gradient and softmax dependent perturbations to the data, slightly improving the performance. More recently, the work in \cite{lee2018simple} assumed that the feature space of the penultimate layer of a classifier follows a Gaussian distribution, allowing for the estimation of mean and variance statistics from the features of every class. A posterior fit of a multivariate Gaussian distribution and the usage of Mahalanobis distance to gauge the closest class-conditional distribution is used for OoD detection. Following the strategy posed in \cite{liang2017odin}, the performance is improved by adding small noise perturbations; this time depending on the Mahalanobis distance rather than softmax. Further rationale for the effectiveness of Mahalanobis distance for OoD detection is given in \cite{kamoi2020mahalanobis}.

Contrarily to previous approaches, in \cite{hendrycks2018deep} a technique called Outlier Exposure is presented for improving the performance of existing detectors. Authors propose to leverage the enormous quantity of data available nowadays by modifying the training process of neural networks with the addition of an additional term to the original loss function. This term depends on the original task (classification, density estimation, etc.) and on the detector in use. This term helps the model to learn heuristics that will improve the detector's performance. Later, an energy-based detector was derived in \cite{liu2020energy} by adapting the concept of \textit{Helmholtz free-energy} to deep neural networks, expressing it in terms of the denominator of the softmax activation. Therefore, each sample is assigned an score based on its logits (its \emph{energy}), which is shown to be higher for OoD samples than for ID instances. This was proven to yield even better detection performance than previous softmax-based scores. Thereupon, the study presented in \cite{lin2021mOoD} improved the computational efficiency at inference time of the energy-based detector by training multiple intermediate classifiers operating at different depths of the trained neural network. The intermediate activations of the neural networks are extracted at certain depths of its internal layered structure depending on its complexity, leveraging the intuition that less complex OoD samples can be detected by using low-level activation statistics of the network. Extending the success of the energy-based detectors, authors in \cite{wang2021can} came up with a new scoring function (\textit{JointEnergy}), which exploits the joint uncertainty across labels by aggregating the label-wise free energy for multiple labels. Recently, \cite{sun2021react} proposed a simple post-hoc method forged as \emph{Rectified Activations} (ReAct), which is applied to the penultimate layer of a network to truncate the activations at inference time. Its effectiveness is supported by the finding that applying the BatchNorm statistics of ID training data to the OoD data at inference time leads to abnormally high activations in all layers, and therefore to the model's output, deteriorating the detection performance. Basically, ReAct limits the activation's value, yielding better detection scores than the previously presented techniques hinging on the logits of the network.

%JAVI
 
\subsection{Spiking neural networks} \label{ssec:SNNsRelated}

This family of neural networks has garnered significant attention since their inception, not only because they can represent more faithfully the inner stimuli of the human brain, but also due to the technical advantages that their particular information processing strategy entails in terms of spatio-temporal pattern learning and energy efficiency. In the brain, information flowing between neurons is conveyed through synapses by trains of short electric pulses called \emph{spikes}, so that neurons receive and accumulate them as potential in their membrane. When a potential threshold is met, the neuron emits an pulse (spike) of its own to the subsequent neuron. The neuron emitting a spike is the pre-synaptic neuron, whereas the receiver neuron is referred to as post-synaptic neuron. Mimicking this dynamic behavior offers an inherent way to deal with temporal data, and the sparsity and binary nature of spikes makes it possible to reduce dramatically the energy consumption of these models when deployed on specialized hardware \cite{merolla2014million}.

A major drawback of SNNs is the fact that spiking neurons are not differentiable due to the discontinuity of the spike. Therefore, gradient backpropagation-based learning schemes are not directly applicable. To overcome this issue, plenty of learning methods have been proposed over the years, ranging from new bio-inspired learning rules to adaptations of the gradient descent algorithm to deal with the aforementioned discontinuity, among others. To top it all off, each learning method comes with its own limitations in terms of input encoding scheme, neuron models, network architectures and network running mode, among others \cite{wang2020supervised}. In this subsection we briefly revisit them as a referential knowledge base for concepts later referred to in the remainder of the article.

Encoding methods for SNNs can be divided into two big groups: rate encoding schemes and temporal encoding schemes \cite{auge2021survey}. In both cases, their purpose is to transform input samples into spike trains that can be processed by the network. As such, rate encoding schemes embeds input data information in the instantaneous or averaged rate of generated spikes in a single or group of neurons, i.e., in the count of spikes emitted within a time encoding window. On the other hand, temporal encoding methods conveys the input information in the precise timing of the spikes. 

As for the neuron models, a great variety exists, each with its own biological plausibility and processing efficiency. Basically, they describe the evolution of the membrane potential of the neuron and the spike generation at different levels of detail. The Leaky Integrate-and-Fire (LIF) model, for instance, is the most implemented model in the SNN community due to its low computational cost and simplicity when describing neuron dynamics \cite{kasabov2019time}. Basically, the neuron integrates the input spikes as currents applied to a resistor–capacitor (RC) circuit that increases its voltage until a threshold is exceeded. At this point in time, a spike is emitted and the voltage of the neuron is reset to the resting value. The LIF neuron abstracts away the shape and profile of the output spike: it is simply treated as a discrete event. As a result, information is not stored within the spike, but rather the timing (or frequency) of spikes. The Spike-Response Model (SRM) behaves similarly to the LIF model, with the main difference being that it has an adjustable threshold that depends on the time since the last post-synaptic spike occurred. While LIF models are usually defined in terms of differential equations, the SRM expresses the membrane potential at a certain time as an integral over the past \cite{gerstner2002spiking}. More biologically plausible models include the Hodking-Huxley or the Izhikevich models \cite{izhikevich2003}, among others. 

As in traditional neural networks, SNNs can run in an offline or online fashion. SNNs have shown their ability to build spike-time learning rules that capture temporal associations by leveraging spike information representation. Therefore, not only they are well suited for learning from spatio-temporal data in a batch learning setting, but they are also inherently effective for handling these temporal associations in streaming data \cite{lobo2020spiking}.

After the above short primer on concepts related to SNNs, we proceed by describing several learning algorithms relevant to our study. It is not the aim of this work to comprehensively review all existing learning methods for SNNs. Therefore, some few insights about the most relevant approaches for this purpose are next provided. To begin with, the Spike-Time Dependent Plasticity (STDP) \cite{caporale2008spike} is a strongly bio-inspired unsupervised learning rule that strengthens the connection between pre- and post-synaptic neurons that fire one after the other and that has inspired many other learning algorithms. SpikeProp \cite{bohte2002error} was the first to train SNNs by backpropagating errors, done on a single hidden layer architecture, with the SRM neuron model and the limitation that neurons were limited to emitting only one spike. Works thereafter refined this method to enable multiple spike firing at desirable times \cite{booij2005gradient}. The main idea of the Spike Pattern Association Neuron (SPAN)~\cite{mohemmed2012span} is to transform the input and the desired output spike trains into analog signals by convolving the spikes with a kernel function. In this manner, the computation of error signals is simplified, allowing for the application of gradient descent to optimize the synaptic weights.

However, learning methods discussed so far have difficulties to effectively train deep (multi-layered) SNNs. Surrogate Gradient (SG) methods overcome the discontinuity of the spiking neuron model by substituting it with a smooth continuous relaxation. Therefore, it does not impose any restrictions in the learning algorithm that can be used to train the model. Consequently, SG methods can be combined with Back-Propagation Through Time (BPTT) to give rise to recurrent flavors of SNNs. Nevertheless, this workaround poses stringent limitations for SNNs in terms of computational and memory resources when deployed over traditional hardware (GPUs). Moreover, its implementation over neuro-morphic hardware may impose locality requirements that can complicate its deployability \cite{neftci2019surrogate}. A remarkable method that solves the latter and utilizes a surrogate gradient approach is \emph{SuperSpike}, a biologically plausible nonlinear Hebbian three-factor rule with individual synaptic eligibility traces \cite{zenke2018superspike}. Basically, the objective of the algorithm is to minimize the van Rossum distance between the desired and the actual spike train through gradient descent. Hence, the derivatives of the spikes with respect to the weights of the network are needed. To make this computation feasible, the spike train is approximated by replacing it with a continuous auxiliary function that depends on the difference between the membrane potential of the LIF neuron and its firing threshold. 

\subsection{Explainability in spiking neural networks} \label{ssec:xAI_in_SNNs}

As in traditional neural networks, SNNs are inherently opaque models that need to be complemented with explanations and justifications of their issued predictions, particularly in critical applications involving a non-expert audience. Currently it is widely acknowledged that SNNs are not applied in practical modeling problems as often as their non-spiking counterparts. However, their event-based working procedure is making SNNs gradually replace non-spiking neural architectures in problems and tasks driven by computational efficiency. This progressively increasing prevalence of spiking neural computation approaches makes it of utmost necessity to derive XAI methods specific for this family of models, over as many tasks as possible, so that the time to their acceptance and adoption in real-world settings is effectively narrowed.

Despite this noted relevance, efforts invested so far in this direction are scarce. To begin with, the work in \cite{jeyasothy2019novel} proposed a new knowledge encoding method that enables the attribution of the input features in the classification decision in a Multi-Class Synaptic Efficacy Function based leaky-integrate neuRON (MC-SEFRON) classifier, a model without hidden layers based on the one presented in \cite{jeyasothy2018sefron}. MC-SEFRON leverages a time-varying weight neuron model and trains it by using a modified STDP rule previously designed in preceding contributions. The result is a classifier with great accuracy when tested over the MNIST image classification task, and capable of eliciting explanations in the form of attribution maps for input images. The downside is that explanations are tightly bounded to the crafted model, and therefore are not generalizable to other spiking neuron architectures in more frequent use by the related community. Methods for rule extraction from SNNs are presented in \cite{soltic2010knowledge} and also in \cite{doborjeh2021deep,kumarasinghe2020deep}, where the NeuCube architecture \cite{kasabov2014neucube} is used.

When searching for a more general explainability method, the contribution \cite{kim2021visual} exploited the fact that spikes with short inter-spike intervals highly contribute to the neural decision process. This observation gave rise to the Spike Activation Map (SAM) proposed in this study. For each timestep and neuron, a Neural Contribution Score (NCS) aggregates the Temporal Spike Contribution Score (TSCS) of all the pre-synaptic spikes previously processed by every neuron. These NCS are calculated at every time step for the feature maps of a convolutional neural network, and are summed pixel-wise across all channels, obtaining the aforementioned SAMs. Hence, one heatmap is generated by this technique for every time step and SNN layer.

When shifting the focus on time series data, the contribution in \cite{nguyen2021temporal} combines three components to design a local feature-based explanation called Temporal Spike Attribution (TSA): i) the influence of spike times, modeled by the NCS introduced above; ii) the influence of the model's parameters, extracted from the weights of the connections between layers; and iii) the influence of the classification decision, inferred from the softmax values. To the best of our knowledge, this threefold approach is the more recent of the short series of contributions dealing with explainability techniques specifically devised for SNNs.

\subsection{Contribution over the state of the art} \label{ssec:contribution}

After examining the current status of knowledge of the research fields tackled in our work, we note that multi-layered SNNs are lately becoming a viable option to implement and test in practical settings, performing competitively against non-spiking neural counterparts with notably improved energy efficiency. For instance, spiking-based CNN models comprising several neural processing layers have been proposed recently for different image classification tasks \cite{wu2019direct, samadzadeh2020convolutional, lian2022training, vaila2021deep}. This is the reason why the community is slowly expressing interest in addressing practical issues of these models when used in real-world environments, including the detection of OoD samples. However, as far as our revision of the literature can tell, there is no prior work in which OoD detection is studied for SNNs, nor have explanations been fabricated for modeling tasks besides classification. This work steps beyond the current state of the art by designing a specific method to detect OoD samples for SNNs, validating that not only it can be done by examining information produced during the training process of the SNN, but also that it attains better detection statistics than post-hoc methods from the literature that can be directly applied to these models. In addition, the work proposes an attribution method aimed to make the OoD detection process more understandable and trustworthy for the user, which spans further the narrow gamut of modeling tasks for which XAI methods have so far been developed in related studies.
%JAVI

\section{Proposed Approach} \label{sec:ProposedApproach}

\begin{figure}[ht]
    \centering
    \includegraphics[width=1\textwidth]{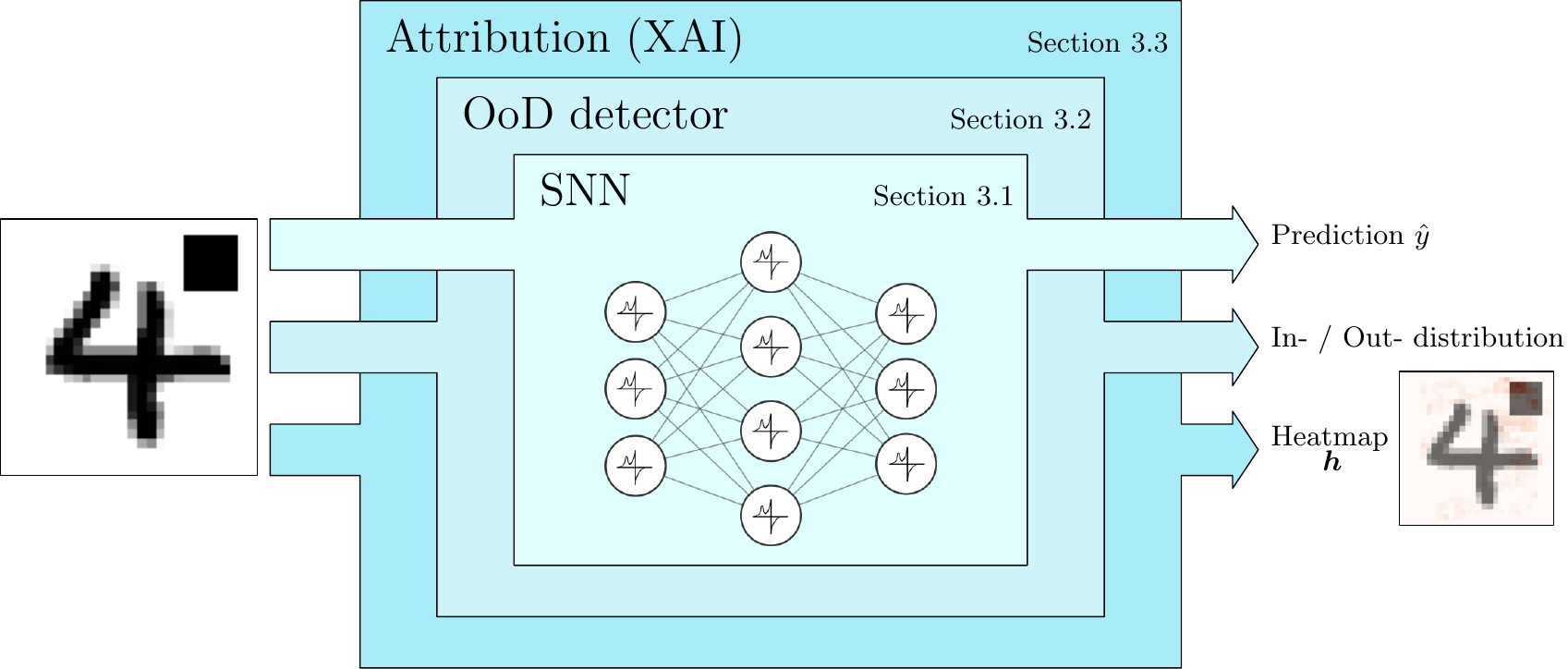}
    \caption{Workflow of the SCP based detector for an image classification task.}
    \label{fig:SCP_Workflow}
\end{figure}

We now describe the method designed to detect OoD samples in SNNs, which we hereafter denote as \textit{Spike Count Pattern} (SCP) based detector. Its workflow is graphically sketched in Figure \ref{fig:SCP_Workflow}. First, a mathematical formulation of the network characteristics is posed in Subsection \ref{ssec:Preamble}. Next, an in-depth explanation of the detector is provided in Subsection \ref{ssec:OoDdetecion}. Finally, Subsection \ref{ssec:Attribution} describes the steps to produce an attribution map based on the information generated during the OoD detection performed by the SCP-based detector.

\subsection{Spiking Neural Network: mathematical notation} \label{ssec:Preamble}

Our method targets the detection of OoD instances by exploiting the internals of SNNs working with rate-based encoding strategies. We assume a training dataset $\mathcal{D}_{tr}$ with samples $\{\mathbf{x}^{(i)}\}_{i=1}^M$, where $\mathbf{x}^{(i)}\in \mathbb{R}^D$, $M$ denotes the number of samples in the training set, and $D$ is the dimension of every input instance. Each of such training samples belong to a class $c\in\{1,2,\ldots,C\}$, where $C$ denotes the number of classes in $\mathcal{D}_{tr}$. The SNN model requires spike trains $\mathbf{s}^{(i)}(t)=\{s_d^{(i)}(t)\}_{d=1}^D$ at their input which, when implemented in discrete-time processing hardware, are represented by binary sequences \smash{$\mathbf{s}^{d,(i)}=\{s_1^{d,(i)},s_2^{d,(i)},\ldots,s_{T/\Delta T}^{d,(i)}\}$}, where \smash{$s_t^{d,(i)}\in\{0,1\}$} and a one represents the occurrence of a spike. Each sample will be processed during a certain amount of time, named the simulation time $T$, that is divided in time steps of width $\Delta T$. The number of time steps within a simulation period is easily obtained as $T/\Delta T$, which establishes the length of the binary sequence in which spikes are represented in the discrete domain.

Hence, inputs must be converted in spike trains by an encoding method $g(\mathbf{x}^{(i)}):~\mathbb{R}^D \mapsto \{0,1\}^{t \times D}$, where the addition of the time dimension is done if the data is static. Here a rate-based coding scheme that converts each feature to a spike train using an homogeneous Poisson process is employed \cite{heeger2000poisson}. Basically, for each feature, the probability of a spike \smash{$s_t^{d,(i)}$} being emitted at time $t\in\{1,\ldots,T/\Delta T\}$ for feature $d\in\{1,\ldots,D\}$ and instance $\mathbf{x}^{(i)}$ is given by:
\begin{equation}
    Pr\{s_t^{d,(i)}\} = r\Delta T = x_d^{(i)} \: r_{max}\:\Delta T,
    \label{eq:Poisson}
\end{equation}
where $\Delta T$ is the interval width, and $r$ denotes the firing rate whose value depends on the value of each normalized feature $x_d^{(i)} \in \mathbb{R}[0,1]$ and a maximum firing rate threshold $r_{max}$. The value of $r_{max}$ is upperbounded by $1/\Delta T$ to avoid generating more than one spike per interval. To generate spike trains, at each timestep a sample $z_{t}^{d,(i)}$ is drawn uniformly at random from a continuous distribution $\mathcal{U}(0,1)$, and is compared to the probability in Expression \eqref{eq:Poisson}, such that a spike is generated if \smash{$z_t^{d,(i)} \leq r \Delta T$}. As a result, the obtained spike train $\mathbf{s}^{(i)}(t)$ comprises randomly distributed spikes, such that the number of spikes in the simulation period $T$ follows a Poisson distribution. 

To process an instance $\mathbf{x}^{(i)}$, the SNN converts each of its features $x^{d,(i)}$ into a spike train $\mathbf{s}^{d,(i)}$. At each time step $t$, the corresponding part of all the spike trains \smash{$\{s_t^{1,(i)},\ldots,s_t^{D,(i)}\}$} are fed to the model. Assuming a Leaky-Integrate-and-Fire neuron model, the state of the LIF neurons of each SNN layer is updated depending on the occurrence of a spike in the previous layer and the previous state of the neuron. If an activation threshold is surpassed, the potential/voltage of the neuron is reset to its resting state value for the next simulation time step $t+1$. It is important to note that spikes affect the next layer's neuron states through the so-called \emph{synaptic current}, which impacts on the membrane voltage of the neuron. This synaptic current generated by a spike depends only on the weight of the connection between the actual neuron (post-synaptic) and the neurons of the previous layer (pre-synaptic).

When the SNN is used for classification tasks, the output layer is not composed by LIF neurons, but rather by leaky integrators matching the number of classes in $\mathcal{D}_{tr}$. This type of neurons only accumulates voltage through time, i.e., it lacks any mechanisms to fire a spike. The maximum voltage levels achieved by these neurons during the simulation period $T$ are interpreted as the logits input to a softmax function, from where the corresponding predicted class is inferred under a \emph{winner-takes-all} decision strategy (as in classical non-spiking neural classifiers).

\subsection{SCP-based Out-of-Distribution detection approach} \label{ssec:OoDdetecion}

\begin{figure}[ht]
    \centering
    \includegraphics[width=0.8\textwidth]{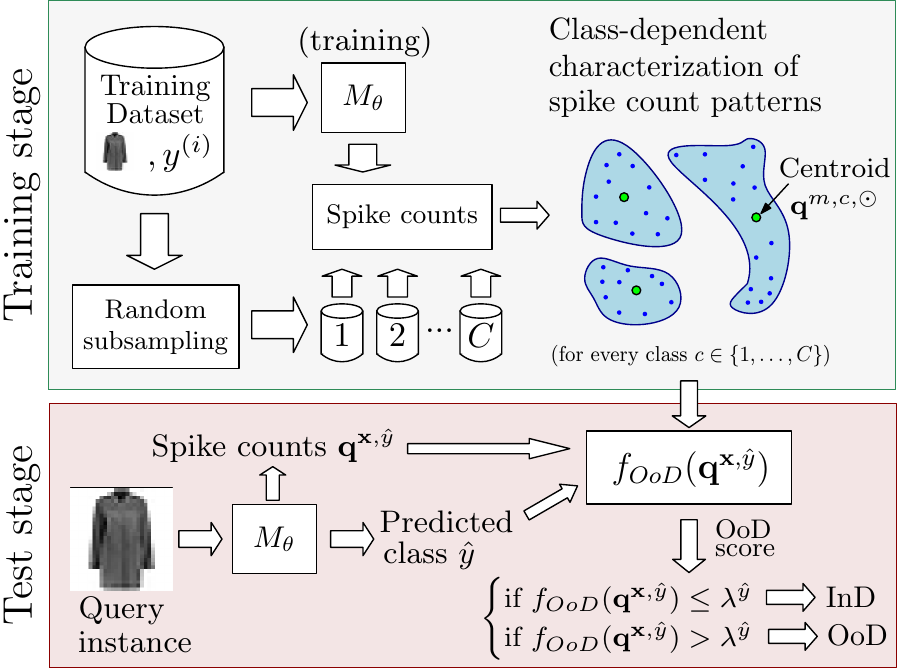}
    \caption{Graphical description of the methodology of the SCP based detector.}
    \label{fig:SCP_Algorithm}
\end{figure}

The design of the proposed OoD detection approach departs from the intuitive observation that similar samples of the same class tend to have similar neuron activation patterns in the hidden layers. Indeed, the SCP-based detector is largely inspired by this observation to yield a novel OoD detection method, which is graphically depicted in Figure \ref{fig:SCP_Algorithm}. In essence, the detection hinges on comparing the spike count pattern of every new test sample to the representative (archetypical) spike count patterns of its predicted label, which can be characterized during the training stage of the SNN. Therefore, the method requires augmenting the training algorithm of the SNN in order to characterize and produce such spike count archetypes, so that they can be used at inference time to detect whether any test instance is out of the training distribution. Here, a spike count $q(\mathbf{s})$ corresponding to a spike train $\mathbf{s}\in\{0,1\}^{T/\Delta T}$ is given by the number of spikes fired during the simulation period, namely, $q(\mathbf{s})=\sum_{t=1}^T s_t$, where we recall that $s_t\in\{0,1\}$ and value $1$ represents a fired spike.

In doing so, and for the sake of a reduced computational complexity added to the training process, the SCP-based detector selects uniformly at random a specific number of training examples of every class and collects the spike counts of the layer prior to the softmax (for reasons disclosed later). Once spike counts have been computed, a class-conditional clustering is performed over the spike counts, using a distance metric to gauge the dissimilarity between spike counts. After computing all pairwise distances, the centroid of each class-conditional cluster resulting from the clustering stage is computed, which can be regarded as the aforementioned  spike count archetypes of the samples inside the cluster of the class to which they belong. It should be noted that every class in $\mathcal{D}_{tr}$ may span more that one cluster, depending on the inherent intra-class variability of the spike counts produced by the training instances of the class at hand. Finally, at inference time, a test instance $\mathbf{x}'$ is processed through the already trained SNN, yielding a predicted class $\hat{y}$. A score is then computed to determine whether a new sample belongs to the in- or out-distribution, which is given by the closest $L_1$ distance to the archetypes of its predicted class $\hat{y}$. If this distance is greater than a predefined threshold $\lambda$, $\mathbf{x}'$ is declared to be an OoD sample. Otherwise, the SCP-based detector determines that $\mathbf{x}'$ belongs to the distribution of the training data.
\begin{algorithm}[htb]
\DontPrintSemicolon
	\KwData{Training data $\mathcal{D}_{tr}$, size $P$ of training samples $\mathcal{D}_{tr}^c$, last layer index $L$, spikes $\mathbf{s}^{n,L,(i)}$ associated to example $\mathbf{x}^{(i)}$, distance measure $d(\cdot)$, clustering algorithm, aggregation function $f_{agg}(\cdot)$, OoD score function $f_{OoD}(\cdot)$, thresholds $\lambda^{\hat{y}}$, query instance~$\mathbf{x}$}
	\KwResult{Decision on the OoD nature of $\mathbf{x}$}
\texttt{//Training stage}\;
	Train SNN over $\mathcal{D}_{tr}$\;
	\For{$c\in\{1,\ldots,C\}$}{
		$\mathcal{D}_{tr}^c \leftarrow$ $P$ sized $c$-class sample  randomly selected from $\mathcal{D}_{tr}$\;
		Collect the spike trains $\mathbf{s}^{n,L,(i)}=\{s_t^{n,L,(i)}\}_{t=1}^{T/\Delta T}$\;
		Compute the spike counts $q(n,L,(i),c)$ as per Expression \eqref{eq:spike_counts}\;
		Aggregate spikes counts into a vector $\mathbf{q}^{(i),c}=\{q(n,L,(i),c)\}_{n=1}^{N_L}$ \;
		Compute distances $d(\mathbf{q}^{(i),c},\mathbf{q}^{(j),c})$ $\forall \mathbf{q}^{(i),c},\mathbf{q}^{(j),c} \in \mathcal{D}_{tr}^c$\;
		Extract $M^c$ clusters $\{\bm{\mathcal{Q}}^{m,c}\}_{m=1}^{M^c}$, where $\bm{\mathcal{Q}}^{m,c}\subseteq \{\mathbf{q}^{(i),c}\}_{i\in\mathcal{D}_{tr}^c}$
		\;
		\For{$m\in\{1,\ldots,M^c\}$}{
		Compute archetype $\mathbf{q}^{m,c,\odot}$ applying $f_{agg}()$ to $\bm{\mathcal{Q}}^{m,c})$;
		}
	}
\texttt{//Detection stage}\;
Let $\hat{y} = M_\theta(\mathbf{x})$ (prediction of query sample)\;
Collect spike trains $\mathbf{s}^{n,L,\mathbf{x}}$\;
Compute spike count $\mathbf{q}^{\mathbf{x},\hat{y}}=\{q(n,L,\mathbf{x},\hat{y})\}_{n=1}^{N_L}$\;
Assign an OoD score by computing $f_{OoD}(\mathbf{q}^{\mathbf{x},\hat{y}})$ as per Expression \eqref{eq:scoreOoD}\;
Declare $\mathbf{x}$ to be OoD if $f_{OoD}(\mathbf{q}^{\mathbf{x},\hat{y}})>\lambda^{\hat{y}}$, otherwise $\mathbf{x}$ follows the in-distribution of $\mathcal{D}_{tr}$\;
\caption{Proposed SCP detector}
\label{alg:SCP}
\end{algorithm}

We proceed by describing step by step the SCP-based detector. As schematically described in Algorithm \ref{alg:SCP}, our developed technique requires a training stage for generating archetypical spike count patterns for every class in the dataset. For simplicity we assume $L$ stacked layers in the SNN architecture, comprising $N_L$ LIF neurons. By slightly modifying the previous notation, we denote the time slot of the spike train occurring at the $n$-th neuron of the last layer $L$ (the layer prior to the leaky integrator) as $\mathbf{s}^{n,L,(i)}=\{s_t^{n,L,(i)}\}_{t=1}^{T/\Delta T}$. Based on this definition, the SCP-based detector starts by extracting $P$ spike trains $\{\mathbf{s}^{1,L,(i)}\}_{i\in\mathcal{D}_{tr}^c}$ for each of the labels in the training dataset (line 5 of Algorithm \ref{alg:SCP}). To this end, a $P$-sized subset $\mathcal{D}_{tr}^c\subset \mathcal{D}_{tr}$ of the training instances predicted to belong to label $c$ is sampled at random (line 4), such that $|\mathcal{D}_{tr}^c|=P$ $\forall c\in\{1,\ldots,C\}$. Once such spike trains are collected, counts can be computed for each spike train at neuron $n$ and instance $\mathbf{x}^{(i)}$ (line 6) as:
\begin{equation}\label{eq:spike_counts}
q(n,L,(i),c)=\sum_{t=1}^{T/\Delta T} s_t^{n,L,(i)},\: \text{for }i\in\mathcal{D}_{tr}^c,
\end{equation}
which can be aggregated into a spike count vector $\mathbf{q}^{(i),c}=\{q(n,L,(i),c)\}_{n=1}^{N_L}$ (line 7). The proposed SCP-based detector extracts the spikes of the \textit{last layer} $L$ of the SNN architecture for two reasons. The first is that the deeper the selected layer is along the hierarchy of neural layers, the higher the level of the features handled is, and the lower the number of spike count trains will be. High-level semantics are better for OoD sample discriminability, whereas low-dimensional spaces are preferred for characterizing the space of archetypical behaviors of the activations of instances of a given class. Second is the fact that the softmax layer is bounded to represent the probability of the input belonging to a certain class, which may not be well suited to characterize OoD samples.

Based on the $P$ spike count vectors $\{\mathbf{q}^{(i),c}\}_{i\in\mathcal{D}_{tr}^c}$ computed for every class $c$ as per Expression \eqref{eq:spike_counts}, a clustering algorithm can be used over such spike counts to group them in terms of their inter-vector similarity (line 9). This clustering process yields $M^c$ clusters $\{\bm{\mathcal{Q}}^{m,c}\}_{m=1}^{M^c}$, where $\bm{\mathcal{Q}}^{m,c}\subseteq \{\mathbf{q}^{(i),c}\}_{i\in\mathcal{D}_{tr}^c}$ denotes the $m$-th cluster arising from the spike count activations of the random subsample drawn from the training instances predicted to belong to class $c$. To compute the dissimilarity between spike count vectors we select the $L_1$ norm (also referred to as \emph{Manhattan distance}) between two vectors (line 8), namely:
\begin{equation} \label{eq:L1dist}
d(\mathbf{q}^{(i),c},\mathbf{q}^{(j),c})=\sum_{n=1}^{N_L} |q(n,L,(i),c)-q(n,L,(j),c)|,
\end{equation}
as its behavior in high-dimensional spaces is known to behave better than other alternatives, e.g., the Euclidean norm \cite{aggarwal2001surprising}. Finally, all samples in every cluster $\bm{\mathcal{Q}}^{m,c}$ are aggregated to obtain an archetype $\mathbf{q}^{m,c,\odot}$ (line 11), which is considered the representative spike count pattern of all samples in the corresponding cluster. We do so by devising a function $\mathbf{q}^{m,c,\odot}=f_{agg}(\bm{\mathcal{Q}}^{m,c})$, which can be chosen to be any aggregation strategy that allows for a variable-size set of spike count patterns in their argument, e.g. the mean, the median or any other statistic alike.

After computing the archetypes of every cluster and class $\mathbf{q}^{m,c,\odot}$, the OoD detection process is ready to process each new query or test sample $\mathbf{x}$ fed to the model. First, its prediction $\hat{y}$ is output by the trained SNN model (line 13). Along the inference process, spike trains $\mathbf{s}^{n,L,\mathbf{x}}$ are collected at the last layer of the network (line 14), from where spike counts vectors $\mathbf{q}^{\mathbf{x},\hat{y}}=\{q(n,L,\mathbf{x},\hat{y})\}_{n=1}^{N_L}$ are computed (line 15). Next, an score is assigned to the test instance by using a scoring function (line 16):
\begin{equation} \label{eq:scoreOoD}
f_{OoD}(\mathbf{q}^{\mathbf{x},\hat{y}})=\min_{m\in\{1,\ldots,M^{\hat{y}}\}} d(\mathbf{q}^{\mathbf{x},\hat{y}},\mathbf{q}^{m,\hat{y},\odot})
\end{equation}
namely, the minimum $L_1$ distance between the spike count pattern $\mathbf{q}^{\mathbf{x},\hat{y}}$ of test instance $\mathbf{x}$ and the archetypes $\mathbf{q}^{m,\hat{y},\odot}$ for $m\in\{1,\ldots,M^{\hat{y}}\}$ computed for its predicted class $\hat{y}$. Finally, the value of $f_{OoD}(\mathbf{q}^{\mathbf{x},\hat{y}})$ is used to decide whether instance $\mathbf{x}$ belongs to the in- or the out-distribution (line 17). If the score is greater than a certain class-conditional threshold $\lambda^{\hat{y}}$, the sample is considered as an out-of-distribution sample, declaring the sample from the in-distribution otherwise:
\begin{equation}\label{eq:OoDdecision}
    \mathbf{x}\text{ belongs to}\begin{cases}
      \text{in-distribution} & \text{if }f_{OoD}(\mathbf{q}^{\mathbf{x},\hat{y}}) \leq \lambda^{\hat{y}}, \\
      \text{out-distribution} & \text{if }f_{OoD}(\mathbf{q}^{\mathbf{x},\hat{y}}) > \lambda^{\hat{y}}.
    \end{cases} 
\end{equation}

\subsection{Local relevance attribution method} \label{ssec:Attribution}

In case the query instance $\mathbf{x}$ is declared not to belong to the in-distribution, the SCP-based detector is complemented by a local attribution-based explanation technique. This explanatory technique is local, as it operates over a given test instance towards generating a relevance attribution vector $\mathbf{h}(\mathbf{x}) \in \mathbb{R}^{D}$, where each component $h_d(\mathbf{x})$ quantifies the importance of each feature of the input when detecting it as an OoD sample. 

\begin{figure}[h!]
    \centering
    \includegraphics[width=1\textwidth]{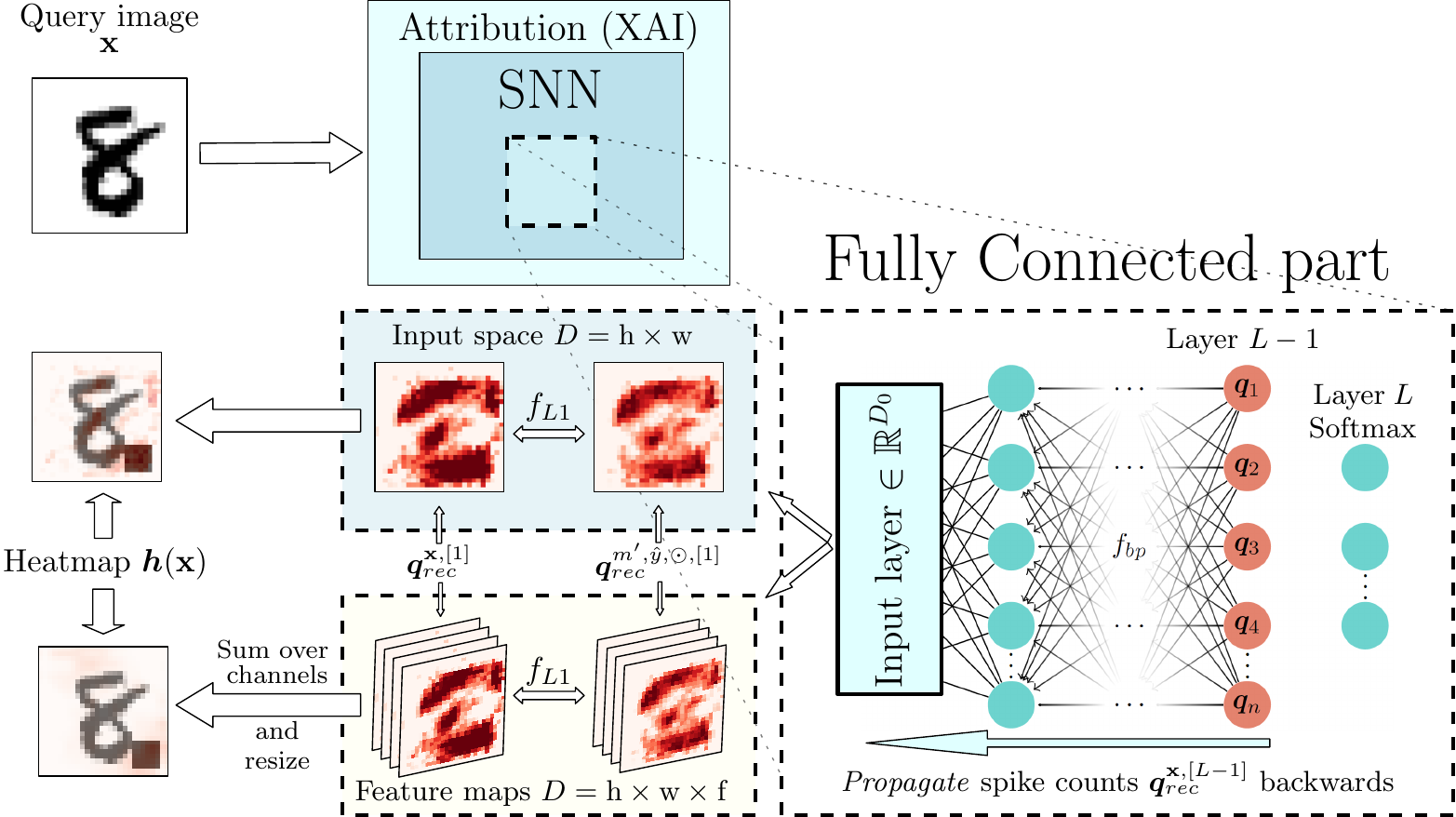}
    \caption{Schematic description of the local attribution method. Here, $\mathrm{h}$, $\mathrm{w}$ and $\mathrm{f}$ refer to height, width and filters respectively.}
    \label{fig:SCP_Atribution_worflow}
\end{figure}

%In this case, $\mathrm{h}$ and $\mathrm{w}$ refer to the input's height and width respectively.
The steps of this local attribution method are schematically illustrated in Figure \ref{fig:SCP_Atribution_worflow} and detailed in Algorithms \ref{alg:backprop} and \ref{alg:attribution}. In the following, we refer as \emph{input layer} to the first fully-connected layer of the SNN network. In the case a feature extractor (e.g. a convolutional layer) is preceding that fully connected stage, the input layer will refer to the output of the feature extractor.

%This way, we expect to obtain an image that will point out the parts of the input that specially drove the detectors decision to classify a sample as an OoD.
%Hence, the former is achieved using the L1 distance like function on the input space, as it is the metric used to assign each sample an OoD score, and the latter is accounted by using the weight matrices $\bm{W}^{[k]}: k\in [1,K-1]$ of the network where $K$ refers to the layer index of the softmax layer. We proceed as follows.
%First, to be able to perform the L1 distance at the input space while accounting for the influence of the input features in the latent space of layer $K-1$, we backpropagate the spike counts of both the input $\bm{q}^{[K-1]}$ and the centroid $\bm{q}_m^{\hat{y},\odot}$ to the input layer using the weight matrices $\bm{W}^{[k]}$ with the help of the $f_{bp}$ function presented in the equation \ref{eq:backpropReLU}:
The core ideas driving the design of this method are two: on one hand, we aim to reproduce the detector's procedure to detect OoD samples in the input space of the SNN rather than in the latent space where the detection is done (namely, the $L$-th layer prior to the softmax); on the other hand, since an attribution is sought, the method must account for the influence of each input feature on the aforementioned latent space. The latter is achieved by \emph{propagating} the spike counts used for OoD detection at layer $L$ back to layer $1$ by using the weight matrices $\bm{W}^{[l]}: l\in [1,L-1]$ of the network; and the former is achieved using a $L_1$ distance over the backpropagated spike counts. Here, $l$ is the layer index. 

The local attribution mechanism proceeds as follows: first, we backpropagate the spike counts of both the input $\bm{q}^{\mathbf{x},\hat{y}}$ and the centroid $\bm{q}^{m,\hat{y},\odot}$ to the input layer using the weight matrices $\bm{W}^{[l]}$. This is accomplished by recurrently applying a function $f_{bp}$ defined as:
\begin{equation} \label{eq:backpropReLU}
    \bm{q}^{\mathbf{x},\hat{y},[l-1]} = f_{bp}(\bm{W}^{[l]}, \bm{q}^{\mathbf{x},\hat{y},[l]}) =  \bm{W}^{[l]^\top}\cdot \bm{q}^{\mathbf{x},\hat{y},[l]},
\end{equation}
where we extend previous notation\footnote{Accordingly, the spike counts $\bm{q}^{\mathbf{x},\hat{y}}$ over which OoD detection is done as per Algorithm \ref{alg:SCP} should be denoted as $\bm{q}^{\mathbf{x},\hat{y},[L]}$.} with superindex $[l]$ to indicate that spike counts correspond to layer $l\in\{1,\ldots,L-1\}$. This function can be viewed as the mapping of the spike counts of one layer to the space of the previous layer. Specifically, the spike counts are multiplied by the transpose $\bm{W}^{[l]^\top}$ of the weight matrix at layer $l$. This operation simultaneously transforms the spike counts at the output of the layer back to its inputs, and implicitly implements a branching process similarly to what occurs in the backward pass of gradient calculus over computational graphs. Therefore, recurrently applying this function over all layers between the penultimate layer ($L-1$) and the input layer ($l=1$), we end up with the spike counts mapped to the input space, obtaining what we have called the \textit{reconstruction} of the spike counts in the input space. This operation is described in Algorithm \ref{alg:backprop}, which performs this mapping. 
\begin{algorithm}[h!]
\DontPrintSemicolon
    \KwData{Weight matrices $\bm{W}^{[l]}$, spike counts $\bm{q}^{\mathbf{x},\hat{y},[L]}$, function to map spike counts to previous layer $f_{bp}(.)$}
    \KwResult{Spike counts mapped to input space $\bm{q}_{rec}^{\mathbf{x},[1]}\in\mathbb{R}^D$}
    $\bm{q}_{rec}^{\mathbf{x},[L-1]} = f_{bp}(\bm{W}^{[L]}, \bm{q}^{\mathbf{x},\hat{y},[L]})$ \;
    \For{$l\in\{L-2,\ldots,1\}$}{
    $\bm{q}_{rec}^{\mathbf{x},[l]} = f_{bp}(\bm{W}^{[l+1]}, \bm{q}_{rec}^{\mathbf{x},[l+1]})$\;
	}
	Let $\bm{q}_{rec}^{\mathbf{x},[1]}=\{q_{rec,d}^{\mathbf{x},[1]}\}_{d=1}^D$\;
	\For{$d\in\{1,\ldots,D\}$}{
	$q_{rec,d}^{\mathbf{x},[1]} = \max(q_{rec,d}^{\mathbf{x},[1]},0)$\;
	}
\caption{Reconstruction of spike counts at the input layer}
\label{alg:backprop}
\end{algorithm}

It is important to remark the reason behind the maximum or ReLU operation done in the last step of Algorithm \ref{alg:backprop}. To obtain a reasonable attribution, our goal is to reproduce the operating conditions of our detector, but in the input space rather than in the latent space. As LIF neurons are in use (whose output can be either a spike or nothing), when we integrate them in time (computing the spike counts) we will always obtain a positive or a 0 value, even if the weights arriving a neuron are all negative i.e. inhibitory. Therefore, to imitate this behavior when backpropagating static values (counts), negative values obtained from Expression \eqref{eq:backpropReLU} should be set to zero. 
\begin{algorithm}
    \KwData{Weight matrices $\bm{W}^{[l]}$, spike counts $\bm{q}^{\mathbf{x},\hat{y},[L]}$ and the closest centroid $\bm{q}^{m',\hat{y},\odot}$ among $\{\bm{q}^{m,\hat{y},\odot}\}_{m=1}^{M^{\hat{y}}}$ resulting from Algorithm \ref{alg:backprop}}
    \KwResult{Relevance attribution vector $\bm{h}(\mathbf{x}) \in \mathbb{R}^{D}$}
    Use Algorithm \ref{alg:backprop} to obtain $s$\;
    Repeat Algorithm \ref{alg:backprop} to extract the reconstructed centroid  $\bm{q}_{rec}^{m',\hat{y},\odot,[1]}$ from $\bm{\mathcal{Q}}^{m',\hat{y}}$ for the closest centroid $m'\in\{1,\ldots,M^{\hat{y}}\}$ to $\bm{q}^{\mathbf{x},\hat{y},[L]}$\;
    \For{$d\in\{1,\ldots,D\}$}{
    $h_d(\mathbf{x})=|q_{rec,d}^{\mathbf{x},[1]}-{q}_{rec,d}^{m',\hat{y},\odot,[1]}$\;
    }
\caption{Construction of the relevance attribution vector}
\label{alg:attribution}
\end{algorithm}

Second, after obtaining the reconstructions $\bm{q}_{rec}^{\mathbf{x},[1]}$ and $\bm{q}_{rec}^{m',\hat{y},\odot,[1]}$ (where $m'$ stands for the closest centroid to the spike counts corresponding to sample $\mathbf{x}$), we seek how to reproduce the OoD detector's operation in this input space. To this end we resort to a $L_1$ distance as in Expression \eqref{eq:L1dist} so as to compute the absolute difference of each feature of both reconstructions, namely:
\begin{equation}
h_d(\mathbf{x})=\left|q_{rec,d}^{\mathbf{x},[1]}-{q}_{rec,d}^{m',\hat{y},\odot,[1]}\right|,\; \forall d\in\{1,\ldots,D\},
\end{equation}
i.e., the relevance attribution vector results from the application of the same $L_1$-based OoD detection criterion used in the proposed SCP-based approach, yet in the reconstruction of the spike counts in the input space. 

To end with the description of the proposed relevance attribution technique, it is important to note that, depending on how the input data is encoded to the hierarchy of fully-connected LIF neural layers, an additional step would be needed to map $\mathbf{h}(\mathbf{x})$ (namely, a $D$-dimensional vector) to the dimensions of the original data. We exemplify this additional operation by describing two cases:
\begin{itemize}[leftmargin=*]
    \item Let us assume that $\mathbf{x}$ is an image, so that it is input to the network by means of a fully-connected input layer with as many LIF neurons as the number of pixels and channels of the image. In this case, $\mathbf{h}(\mathbf{x})$ should be \emph{unflattened} in the same order as the one followed to serialize the input image into the network.
    \item If a convolutional layer is used to extract features from the input image, the dimension $D$ of the relevance attribution network depends on the size of the convolutional filters and their cardinality. In this case, the rearrangement of the relevance attribution vector operates first by rearranging its components to the shape of the feature maps of the convolutional layer, yielding a three-dimensional tensor-like structure with width and height equal to the size of the convolutional filters, and with depth equal to the number of convolutional filters. Then, we perform a depth-wise aggregation of this tensor. Finally, the obtained map is scaled up by interpolation to generate a relevance attribution heatmap with the size of the original input image. 
\end{itemize}

\section{Experimental setup} \label{sec:Experiments}

In order to examine the performance of the proposed detection and attribution techniques, we design a comprehensive experimental setup using SNNs for image classification, aimed to inform with empirical evidence the responses to the following research questions:
\begin{itemize}[leftmargin=*]
	\item RQ1: Does the proposed SCP-based detector perform competitively when compared to post-hoc OoD detectors from the literature adapted to SNN architectures? 
	\item RQ2: Does the local relevance attribution method yield informative explanations about the features that make a given sample be detected as OoD? Which are the limitations of this explainability technique?
\end{itemize}

The design of the experiments devised to answer the above questions include the definition of the SNN architectures (Subsection \ref{ssec:snn_archs}), the in-distribution and out-distribution datasets under consideration (Subsection \ref{ssec:datasets}), the OoD methods selected for comparison and the scores used to measure their performance (Subsection \ref{ssec:scores_comp}), and the methodology followed to qualitatively validate the relevance attribution vectors generated for OoD samples by the proposed SCP-detector (Subsection \ref{ssec:experimental_attribution}). In what follows, details about these design choices are given. All scripts and notebooks producing the results reported in this manuscript have been made publicly available in a GitHub repository (\url{https://github.com/aitor-martinez-seras/OoD_on_SNNs}).

\subsection{Considered SNN architectures}\label{ssec:snn_archs}

We start describing the experimental setup by the specific implementation of the SNN under consideration. SNNs comprise a very wide research field that lies in between Neuroscience and ML. Hence, some SNN variants are designed departing from the bio-plausibility perspective, attempting to make a machine simulate faithfully the biological brain, neurons and its synapses. Other variants, however, pursue SNNs capable of solving complex modeling tasks due to their inherent advantages rather than the biological plausibility of the dynamics and neurons used. The latter design criterion has dominated this area in recent times due to the need for more efficient ML algorithms. As a result, a great variety of SNN frameworks are nowadays available \cite{garcia2021preliminary}. Although some of them seek a realistic simulation of the human brain, others offer simpler albeit more effective SNN implementations for data-based modeling, in terms of the the depth of the networks that can be trained and the complexity of the tasks they can tackle. 

In our experiments we focus on practical SNN implementations. Specifically, we use $2^{nd}$ order LIF neurons that account for synaptic conductance. By denoting the membrane voltage as $\vartheta$ and the synaptic current as $\zeta$, these decay over time at rates $\tau_{mem}$ and $\tau_{syn}$, respectively. The ordinary differential equations that are integrated over time (in discrete time steps) are \cite{gerstner2014neuronal}: 
\begin{align}\label{eq:LIF}
    \dot{\vartheta} &= 1/\tau_{\text{mem}} (\vartheta_{\text{leak}} - \vartheta + \zeta), \\
    \dot{\zeta} &= -1/\tau_{\text{syn}} \zeta,
\end{align}
together with the jump condition:
\begin{equation}\label{eq:threshold_act}
    z = \Theta(\vartheta - \vartheta_{\text{th}}),
\end{equation}
and transition equations:
\begin{align}\label{eq:current_update}
    \vartheta &= (1-z) \vartheta + z\cdot \vartheta_{\text{reset}}, \\
    \zeta &= \zeta + \zeta_{\text{in}},
\end{align}
where $\vartheta_{\text{th}}$ defines the membrane voltage that, when exceeded, leads to the emission of a spike; $z$ is a variable taking value 1 if the neuron voltage surpasses $\vartheta_{\text{th}}$ (0 otherwise); $\vartheta_{\text{leak}}$ refers to a constant voltage value that is leaked from the neuron; $\Theta(\cdot)$ stands for the Heaviside step function that returns value 1 when the threshold is surpassed; and $\vartheta_{\text{reset}}$ is the value of voltage that is set when the latter happens. A SNN framework that features this LIF model is Norse \cite{pehle2021norse}, which runs on top of PyTorch and provides rate-based encoding strategies, LIF-based and Izhikevich neuron models, and the SuperSpike training algorithm mentioned in Section \ref{ssec:SNNsRelated}. 

We have tested our method over two different SNN architectures for image classification: a Fully Connected (FC) architecture and a Convolutional Neural Network (CNN) architecture. In both cases, the encoder used to translate pixel values to spike trains is the so-called Poisson encoder, whereas neuron models are LIF (intermediate neurons) except for the output layer, which utilizes Leaky Integrators (LI) followed by SoftMax. LI cells are similar to LIF neurons, but never fire: they only accumulate voltage. Parameters for both types of cells are $\vartheta_{\text{leak}} = 0$, $\vartheta_{\text{reset}} = 0$, $\tau_{\text{mem}} = 0.005 \text{ ms}^{-1}$, and $\tau_\text{{syn}} = 0.01 \text{ ms}^{-1}$. These parameters are identical for all cells in the SNN, except for the voltage threshold $\vartheta_{\text{th}}$, which is specified layer-wise in Table \ref{tb:Architectures} where the architectures are described. The table can be interpreted as indicated in its caption.
\begin{table}[!htbp]
\caption{SNN architectures considered in the experiments. Each layer is defined by a number and some letters, which respectively define the number of neurons and the type of connections. Each layer is separated by a hyphen. $\mathbf{E}$ denotes Poisson encoding neurons, \textbf{fc} stands for fully connected neurons, and \textbf{conv} means convolutional filters. The subscript of the last two defines the voltage threshold $\vartheta_{\text{th}}$ of the LIF neurons. In the last layer only leaky integrators (LI) are used, followed by a SoftMax activation, being the number of output neurons equal to the number of classes in the in-distribution dataset.} \label{tb:Architectures}
\centering
\renewcommand\arraystretch{1.25} 
\resizebox{\textwidth}{!}{\begin{tabular}{ccl}
    \toprule
    Network Type & Reference & \multicolumn{1}{c}{Architectures} \\
    \midrule
    \multirowcell{2}{Fully Connected SNN\\(FC-SNN)} & $FC_1$ & $784\mathbf{E} - 200\textbf{fc}_{0.25} - \bm{LI}$\\
    & $FC_2$ & $784\mathbf{E} - 300\textbf{fc}_{0.25} - 200\textbf{fc}_{0.25} - \bm{LI}$ \\
    \midrule
    \multirowcell{2}{Convolutional SNN \\ (CNN-SNN)} & $CNN_1$ & $784\mathbf{E} - 20\textbf{conv}_{0.2}\textbf{avgpool} - 50\textbf{conv}_{0.2} - \textbf{Flatten} - 300 \textbf{fc}_{0.1} - \bm{LI}$\\
    & $CNN_2$  & $784\mathbf{E} - 20\textbf{conv}_{0.2}\textbf{avgpool} - 50\textbf{conv}_{0.2} - \textbf{Flatten} - 500 \textbf{fc}_{0.1} - 300 \textbf{fc}_{0.05} -\bm{LI}$\\
    \bottomrule
    \end{tabular}}
 \end{table}

All models are trained for $5$ epochs using an Adam optimizer with learning rate equal to $0.002$. The simulation time $T$ is $50$ ms, whereas the time step duration $\Delta T$ is $1$ ms, hence yielding $50$ time steps per sample.

\subsection{In-distribution and out-distribution datasets}\label{ssec:datasets}

In order to evaluate the performance of the proposed SCP-based detector, we have followed common practice in the literature related to OoD detection. Consequently, datasets of well-known image classification tasks have been considered as in-distribution datasets. Specifically, we have used datasets of $28\times 28$ grayscale centered images: the \texttt{MNIST} dataset \cite{deng2012mnist}, composed by $60,000$ images of handwritten images with 10 classes (digits from 0 to 9); \texttt{Fashion MNIST} \cite{xiao2017fashion}, which comprises $60,000$ images of 10 different classes of clothes; \texttt{KMNIST} \cite{clanuwat2018kmnist}, consisting of $60,000$ images of Japanese characters (\emph{Kanjis}), each belonging to one among $10$ classes; and finally, EMNIST \texttt{Letters} \cite{cohen2017emnist}, which provides $145,000$ images of $26$ types of letters of the Latin alphabet.

When it comes to out-of-distribution datasets, we consider all possible combinations of pairs of the in-distribution datasets defined above, establishing one of them as the dataset over which the SNN is trained, and querying the trained SNN with the other. For the sake of a better experimental coverage, we add a few more out-of-distribution datasets, namely: \texttt{notMNIST} \cite{bulatov2011notmnist}, which gathers font glyphs extracted for the letters A through J; \texttt{omniglot} \cite{lake2015omniglot}, which amounts to $1,623$ different handwritten characters from $50$ different alphabets; and \texttt{CIFAR10-BW}, the black-and-white re-scaled version (to $28\times 28$ pixels) of the \texttt{CIFAR10} dataset \cite{krizhevsky2009cifar}. As a result, the SNN trained for every in-distribution dataset will be queried with images drawn from $6$ different out-of-distribution datasets. In all cases, the OoD dataset is reduced to a uniform stratified sample of $10,000$ samples. 

\subsection{Methods for comparison and performance scores}\label{ssec:scores_comp}

For comparison purposes, to the best of our knowledge there is unfortunately no OoD detector specifically designed for SNNs. Consequently, our benchmark considers post-hoc OoD techniques that essentially rely on the output of the neural network being monitored, either the logits or the output of the softmax activation. In the case of the SNN architectures selected for our experimentation, logits correspond to the maximum voltage reach by every LI cell during the simulation time. The methods against which the SCP-based detector is compared include the Baseline method \cite{hendrycks2016baseline}, ODIN \cite{liang2017odin} and the energy-based approach presented in \cite{liu2020energy}. The Baseline method is does not require any parameter tuning, whereas ODIN and the energy-based method are governed by several parameters such as the temperature or the input preprocessing (in the case of ODIN). Values of these parameters must be properly set in order to obtain gOoD results. To this end, although the study in \cite{liu2020energy} states that the energy-based detector can be used parameter-free by defining $T = 1$, we have evaluated several temperature values for each in- vs out-distribution setting, reporting on the configuration that yields the best performance in each case. A similar configuration strategy has been followed for the temperature parameter of ODIN, whereas for the input preprocessing no perturbation is used, following the methodology of the original paper in which this OoD detector was proposed \cite{liang2017odin}. 

Regarding the configuration of the proposed SCP-based detector, the subset of instances of each class used for the characterization of the spike count patterns is formed by $P=|\mathcal{D}_{tr}^c|=1,000$ samples per class. Agglomerative hierarchical clustering was used to extract clusters from the spike counts of all training instances used for characterization. As for the creation of the centroids, a median statistic was selected as the aggregation function $f_{agg}(\cdot)$, which is robust against spike count outliers. Finally, the threshold for each class $\lambda^{\ell}$ is selected utilizing a subset of $1,000$ samples per class extracted from the training data and different from the ones used for the creation of the centroids. To do so, the OoD scores are computed and the threshold $\lambda^{c}$ is defined as the value that yields a target percentage of that class-wise subset that will be correctly classified as in-distribution. 

To quantify the capability of the counterparts in the benchmark to detect OoD samples, several scores used in the literature related to OoD detection are measured:
\begin{itemize}[leftmargin=*]
\item FPR95 (False Positive Rate at 95\% True Positive Rate), that is, the rate of OoD samples wrongly classified as positive (in-distribution) when the True Positive Rate (TPR) is 95\%. Here, TPR and FPR are defined in the usual way for binary classification, i.e., $\text{TPR} = \text{TP}/(\text{TP}+\text{FN})$ and $\text{FPR}=\text{FP}/(\text{FP}+\text{TN})$. In what follows a TP refers to an in-distribution sample correctly classified as such, whereas a TN stands for an OoD sample detected correctly by the OoD detector at hand.
\item AUROC (\emph{Area Under the Receiver Operation Characteristic curve}), which is a threshold-independent score for binary classification that can be regarded as the probability that the model ranks a random positive example with higher score than a random negative example. It is defined as $\text{TPR}/\text{FPR}$.
\item AUPR (\emph{Area Under the Precision-Recall curve}), another threshold-invariant score closely related to AUROC. The precision-recall curve is given by the relationship between $\text{TP}/(\text{TP}+\text{FP})$ (\emph{precision}) and TPR (\emph{recall}). 
\end{itemize}

\subsection{Evaluating the quality of the relevance attribution method}
\label{ssec:experimental_attribution}

When aiming to inform an answer to \textit{RQ2}, there is no standardized way to measure the quality of explanations provided by a relevance attribution technique like the one proposed in this manuscript. To overcome this, we opt for evaluating our method against out-distribution datasets whose instances are built from in-distribution samples with imprinted visual artifacts. Since such artifacts are a priori known to be the parts of the image that \emph{push most} the input sample towards the out-distribution, the relevance attribution technique should highlight such artifacts in the produced heatmap for the image at hand. In other words, if the difference between two samples is a certain extra and/or missing feature, our relevance attribution method should focus on them when performing OoD detection, and therefore should be clearly distinguishable in the heatmap.

\begin{figure}[htb]
    \centering
    \includegraphics[width=0.65\textwidth]{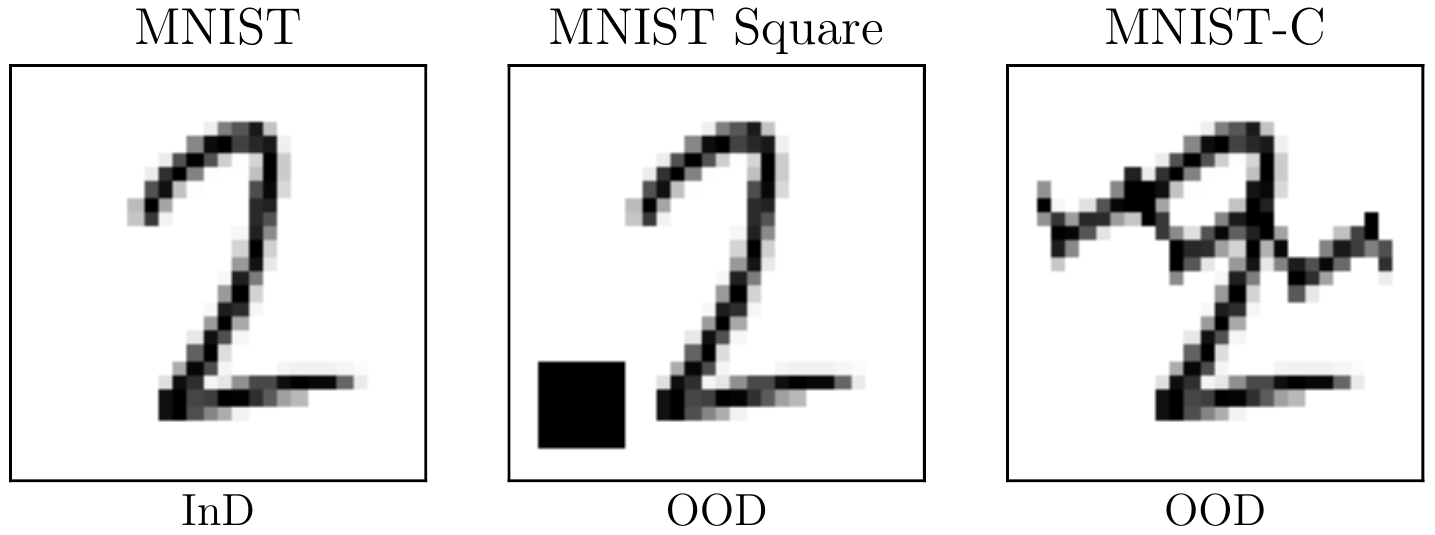}
    \caption{Datasets for the experiments related to the quality of relevance attribution heatmaps.}
    \label{fig:attribution_datasets}
\end{figure}

In doing so, \texttt{MNIST} will be used as in-distribution dataset, whereas out-of-distribution datasets used for the qualitative attribution assessment include i) \texttt{MNIST-Square}, that is, the \texttt{MNIST} dataset with a square of $5 \times 5$ pixels placed randomly in one of the corners of the image (with a separation to the border of the image of 2 pixels); and ii) \texttt{MNIST-C} \cite{mu2019mnist}, another \texttt{MNIST}-based dataset that includes several modifications made to the digits. Specifically, we use the \textit{zig-zag} corruption pattern in our experiments. An example showing the three datasets used in our experiments related to RQ2 is depicted in Figure \ref{fig:attribution_datasets}.

To correctly visualize the heatmaps, an appropriate plot range must be selected previously. To that end, a random sample is extracted from the training set and its heatmaps are computed. Then, from the pixel values of all the heatmaps, the maximum value and the $q$-th quantile value are obtained, where $q$ is the TPR selected to define the threshold. The maximum of the visualization range will be the calculated maximum value and the minimum will be the $q$-th quantile value. The size of the random sample used to compute these maximum and quantile values is set to 100 examples.

\section{Results and discussion} \label{sec:Results}

This section presents the results obtained from the experiments, along with an analysis that leads to answers to the previously introduced research questions: RQ1 (Subsection \ref{ssec:OoDperformance}) and RQ2 (Subsection \ref{ssec:OoDattribution}). The latter includes a short analysis of the limitations that the elicited relevance attribution maps undergo in regards to the complexity of the SNN network structure, as well as the visual features observed by the trained SNN.

\subsection{RQ1: Does the proposed SCP-based detector perform competitively when compared to post-hoc OoD detectors from the literature adapted to SNN architectures?}
\label{ssec:OoDperformance}

Scores obtained to address \textit{RQ1} are shown in Tables \ref{tb:FCresults} (for the FC-SNN architecture) and \ref{tb:CONVresults} (for the CNN-SNN architecture). In both tables, dark gray shaded cells indicate, for each dataset configuration and score, the winning approach between the four methods under comparison. Likewise, light gray shaded cells correspond to the second best method in terms of AUROC score (in general the ranking of the methods in the benchmark as per this score is maintained for the rest of scores).
\begin{table}[!htbp]
\caption{OoD detection scores obtained for the FC-SNN architecture. The arrow next to every score indicates whether higher ($\uparrow$) or lower ($\downarrow$) values correspond to a better detection performance.}
\label{tb:FCresults}
\centering
\renewcommand\arraystretch{0.85} 
\resizebox{\textwidth}{!}{\begin{tabular}{cccccccccccccccc}
    & & \multicolumn{4}{c}{AUROC $\uparrow$} & & \multicolumn{4}{c}{AUPR $\uparrow$} & & \multicolumn{4}{c}{FPR95 $\downarrow$} \\
    \cmidrule{3-6} \cmidrule{8-11} \cmidrule{13-16}
    \makecell{ID\\dataset}& \makecell{OoD\\dataset} & \makecell{SCP\\(proposed)} & Baseline & ODIN & Energy & &   \makecell{SCP\\(proposed)} & Baseline & ODIN & Energy & &  \makecell{SCP\\(proposed)} & Baseline & ODIN & Energy  \\
    \midrule
    \multirow{6}{*}{\texttt{MNIST}}
    & \multicolumn{1}{l}{\texttt{FMNIST}}     &  \mejor 91.21 & 73.97 & \seg 73.97 & 39.51 & & \mejor 87.15 & 65.29 & 65.29 & 41.22 & & \mejor 24.95 & 77.11 &  77.11 & 92.91\\
    & \multicolumn{1}{l}{\texttt{KMNIST}}     &  \mejor 94.89 & 80.75 & \seg 80.75 & 60.62 & & \mejor 94.92 & 76.27 & 76.27 & 55.12 & & \mejor 24.99 & 72.87 &  72.87 & 88.48\\
    & \multicolumn{1}{l}{\texttt{Letters}}    &  \mejor 88.11 & 81.43 & \seg 81.43 & 65.60 & & \mejor 79.49 & 64.19 & 64.19 & 43.04 & & \mejor 51.66 & 73.89 &  73.89 & 89.35\\
    & \multicolumn{1}{l}{\texttt{notMNIST}}    &  \mejor 99.14 & 64.45 & \seg 64.45 & 21.34 & & \mejor 99.01 & 57.16 & 57.16 & 34.57 & & \mejor 03.60 & 86.65 &  86.65 & 97.13\\
    & \multicolumn{1}{l}{\texttt{omniglot}}   &  71.56 & 92.70 & \mejor 94.72 & \seg 94.55 & & 71.90 & 92.79 & \mejor 94.28 & 94.22 & & 88.93 & 46.08 & \mejor 29.08 & 28.34\\
    & \multicolumn{1}{l}{\texttt{CIFAR10-BW}} &  \mejor 99.64 & 80.03 & \seg 80.03 & 17.00 & & \mejor 99.55 & 78.82 & 78.82 & 33.37 & & \mejor 01.36 & 80.59 & 80.59 & 99.67\\
    \midrule
    \multirow{6}{*}{\texttt{FMNIST}}
    & \multicolumn{1}{l}{\texttt{MNIST}}      & 92.60 & 86.82 & \seg 95.98 & \mejor 97.33 & & 93.18 & 85.63 & 95.74 & \mejor 97.65 & & 35.48 & 45.34 & 17.37 & \mejor 15.32\\
    & \multicolumn{1}{l}{\texttt{KMNIST}}     & 84.56 & 78.78 & \mejor 88.36 & \seg 87.26 & & 82.41 & 81.62 & \mejor 88.84 & 87.82 & & \mejor 53.10 & 78.14 & 57.39 & 65.20\\
    & \multicolumn{1}{l}{\texttt{Letters}}    & \mejor 92.83 & 81.42 & 91.83 & \seg 92.23 & & 88.10 & 68.08 & 85.15 & \mejor 88.60 & & \mejor 34.49 & 63.33 & 39.82 & 52.01\\
    & \multicolumn{1}{l}{\texttt{notMNIST}}    & \mejor 98.87 & 61.70 & \seg 61.70 & 43.61 & & \mejor 99.14 & 61.75 & 61.75 & 44.11 & & \mejor 02.47 & 92.33 & 92.33 & 95.78\\
    & \multicolumn{1}{l}{\texttt{omniglot}}   & 84.75 & 91.35 & \seg 97.86 & \mejor 98.48 & & 85.37 & 90.84 & 97.83 & \mejor 98.77 & & 63.55 & 33.90 & 09.97 & \mejor 06.20\\
    & \multicolumn{1}{l}{\texttt{CIFAR10-BW}} & \mejor 96.83 & 55.00 & \seg 55.00 & 33.19 & & \mejor 97.34 & 59.85 & 59.85 & 40.20 & & \mejor 16.81 & 94.66 & 94.66 & 99.21\\
    
    \midrule
    \multirow{6}{*}{\texttt{KMNIST}}
    & \multicolumn{1}{l}{\texttt{MNIST}}      & 61.85 & \seg 84.32 & \mejor 85.79 & 82.43 & & 61.64 & 87.48 & \mejor 88.57 & 85.79 & & 89.34 & 58.20 & \mejor 51.64 & 63.40\\
    & \multicolumn{1}{l}{\texttt{FMNIST}}     & 75.00 & \seg 76.73 & \mejor 76.73 & 46.91 & & 72.85 & 78.58 & \mejor 78.58 & 49.62 & & \mejor 66.14 & 70.67 & 70.67 & 90.86\\
    & \multicolumn{1}{l}{\texttt{Letters}}    & 78.34 & \seg 84.20 & \mejor 85.14 & 80.67 & & 64.47 & 79.81 & \mejor 80.65 & 74.50 & & 71.49 & 58.54 & \mejor 53.51 & 66.38\\
    & \multicolumn{1}{l}{\texttt{notMNIST}}    & \mejor 96.88 & 67.24 & \seg 67.24 & 38.72 & & \mejor 96.68 & 67.59 & 67.59 & 43.88 & & \mejor 09.52 & 81.37 & 81.37 & 95.75\\
    & \multicolumn{1}{l}{\texttt{omniglot}}   & 58.54 & 93.73 & \mejor 96.43 & \seg 96.16 & & 59.68 & 94.98 & \mejor 97.25 & 97.05 & & 92.58 & 22.34 & \mejor 08.09 & 09.68\\
    & \multicolumn{1}{l}{\texttt{CIFAR10-BW}} & \mejor 95.57 & 79.95 & \seg 79.95 & 45.76 & & \mejor 95.78 & 82.46 & 82.46 & 55.17 & & \mejor 15.77 & 65.41 & 65.41 & 99.47\\
    \midrule
    \multirow{6}{*}{\texttt{Letters}}
    & \multicolumn{1}{l}{\texttt{MNIST}}      & 69.06 & 78.19 & \seg 80.98 & \mejor 81.03 & & 82.30 & 86.92 & 88.73 & \mejor 88.96 & & 88.83 & 75.36 & \mejor 68.13 & 68.53\\
    & \multicolumn{1}{l}{\texttt{FMNIST}}     & \mejor 86.33 & 71.30 & \seg 71.30 & 34.31 & & \mejor 92.08 & 83.66 & 83.66 & 56.42 & & \mejor 47.30 & 88.04 & 88.04 & 93.11\\
    & \multicolumn{1}{l}{\texttt{KMNIST}}     & \mejor 86.31 & 76.41 & \seg 77.60 & 66.14 & & \mejor 92.15 & 85.47 & 85.47 & 76.84 & & \mejor 51.41 & 77.88 & 77.88 & 82.53\\
    & \multicolumn{1}{l}{\texttt{notMNIST}}    & \mejor 98.63 & 58.08 & \seg 58.08 & 25.30 & & \mejor 98.90 & 69.65 & 69.65 & 52.35 & & \mejor 07.48 & 93.94 & 93.94 & 97.79\\
    & \multicolumn{1}{l}{\texttt{omniglot}}   & 77.60 & 94.03 & \seg 97.11 & \mejor 97.30 & & 88.04 & 97.06 & 98.54 & \mejor 98.61 & & 87.11 & 32.72 & 15.54 & \mejor 13.01\\
    & \multicolumn{1}{l}{\texttt{CIFAR10-BW}} & \mejor 99.12 & 52.88 & \seg 52.88 & 07.85 & & \mejor 99.54 & 70.18 & 70.18 & 48.12 & & \mejor 03.68 & 98.02 & 98.02 & 99.86\\
    
    \bottomrule
    \end{tabular}}
 \end{table}

%Before diving deep into the tables, a disclaimer is needed, as the reader may notice that sometimes ODIN and baseline methods obtain same scores. The reason is that we have implemented ODIN such that we have tried several temperature parameters in each setting and we show the best peforming result, that 

Several observations can be done by examining the results corresponding to the FC-SNN model (Table \ref{tb:FCresults}). First, the proposed SCP-based OoD detection method achieves in general better detection scores, specially for the \texttt{MNIST} and \texttt{Letters} datasets. Only results obtained for \texttt{omniglot} are worse for the SCP-based approach than for the rest of methods in the benchmark. In the case of the \texttt{KMNIST} dataset, the SCP-based detector dominates the comparison only when \texttt{notMNIST} and \texttt{CIFAR10-BW} are used as out-distribution datasets. Another interesting fact is that the scenarios where our devised method achieves the best results, all other logit-based OoD detectors fall apart, scoring significantly worse. This suggests that logits in SNNs are unsuitable for OoD detection. Conversely, when any logit-based method performs better than our method, our proposal is never ranked second within the benchmark.

Another observation is that there are some cases where ODIN and Baseline methods perform equally. The reason is that, as explained in Subsection \ref{ssec:scores_comp}, several values of the temperature parameter for each combination of datasets have been tested, reporting in all cases the best score. This could occur for a temperature value equal to 1, which would effectively convert ODIN into the Baseline method. Actually, this occurs in the cases where our method clearly dominates the benchmark, supporting the idea that logits are not suitable for detecting OoD samples in some cases.
\begin{table}[!htbp]
\caption{OoD detection scores obtained for the CNN-SNN architecture.}
\label{tb:CONVresults}
\centering
\renewcommand\arraystretch{0.85}
\resizebox{\textwidth}{!}{\begin{tabular}{cccccccccccccccc}
    & & \multicolumn{4}{c}{AUROC $\uparrow$} & & \multicolumn{4}{c}{AUPR $\uparrow$} & & \multicolumn{4}{c}{FPR95 $\downarrow$} \\
    \cmidrule{3-6} \cmidrule{8-11} \cmidrule{13-16}
    \makecell{ID\\dataset}& \makecell{OoD\\dataset} & \makecell{SCP\\(proposed)} & Baseline & ODIN & Energy & &   \makecell{SCP\\(proposed)} & Baseline & ODIN & Energy & &  \makecell{SCP\\(proposed)} & Baseline & ODIN & Energy  \\
    \midrule
    \multirow{6}{*}{\texttt{MNIST}}
    & \multicolumn{1}{l}{\texttt{FMNIST}}     & \seg 95.22 & 93.82 & \mejor 95.79 & \seg 95.22 & & \mejor 96.20 & 94.44 & 95.90 & 95.42 & & 30.87 & 38.12 & \mejor 22.25 & 27.15\\
    & \multicolumn{1}{l}{\texttt{KMNIST}}     &  \mejor 97.95 & 92.18 & \seg 93.07 & 91.77 & & \mejor 98.26 & 92.61 & 93.15 & 91.79 & & \mejor 09.98 & 44.26 &  36.89 & 43.93\\
    & \multicolumn{1}{l}{\texttt{Letters}}    &  \mejor 91.32 & 86.17 & \seg 87.22 & 85.49 & & \mejor 80.56 & 74.16 & 76.13 & 73.29 & & \mejor 30.53 & 57.58 &  53.12 & 60.35\\
    & \multicolumn{1}{l}{\texttt{notMNIST}}    &  \mejor 98.25 & 89.81 & \seg 91.17 & 90.81 & & \mejor 98.47 & 89.05 & 89.46 & 89.42 & & \mejor 07.81 & 48.88 &  42.87 & 47.54\\
    & \multicolumn{1}{l}{\texttt{omniglot}}   &  \mejor 97.54 & 91.27 & 93.24 & \seg 93.27 & & \mejor 97.91 & 90.78 & 92.47 & 92.54 & & \mejor 11.47 & 45.19 &  33.31 & 33.94\\
    & \multicolumn{1}{l}{\texttt{CIFAR10-BW}} & \seg 96.30 & 95.89 & \mejor 97.48 & 88.98 & & 97.38 & 96.74 & \mejor 97.80 & 97.02 & & 27.13 & 30.25 & \mejor 13.68 & 23.28\\
    \midrule
    \multirow{6}{*}{\texttt{FMNIST}}
    & \multicolumn{1}{l}{\texttt{MNIST}}      &  \seg 97.98 & 71.57 & 87.77 & \mejor 97.46 & & 97.19 & 72.29 & 87.02 & \mejor 97.75 & & \mejor 10.84 & 80.56 &  46.11 & 12.78\\
    & \multicolumn{1}{l}{\texttt{KMNIST}}     &  \mejor 97.06 & 78.28 & 89.85 & \seg 94.23 & & \mejor 97.65 & 80.63 & 90.16 & 94.09 & & \mejor 17.96 & 76.92 &  46.99 & 27.03\\
    & \multicolumn{1}{l}{\texttt{Letters}}    &  \mejor 97.13 & 71.92 & 84.96 & \seg 93.06 & & \mejor 96.01 & 54.40 & 71.64 & 83.01 & & \mejor 16.75 & 78.74 &  52.51 & 31.31\\
    & \multicolumn{1}{l}{\texttt{notMNIST}}    &  \mejor 96.00 & 74.33 & 82.00 & \seg 82.29 & & \mejor 95.65 & 74.21 & 78.61 & 79.01 & & \mejor 19.57 & 79.52 &  66.37 & 70.78\\
    & \multicolumn{1}{l}{\texttt{omniglot}}   &  \mejor 98.01 & 71.26 & 86.46 & \seg 94.15 & & \mejor 98.35 & 70.86 & 84.59 & 94.40 & & \mejor 08.47 & 81.82 &  51.68 & 27.51\\
    & \multicolumn{1}{l}{\texttt{CIFAR10-BW}} & \seg 91.18 & 87.78 & \mejor 94.70 & 89.78 & & 91.82 & 89.48 & \mejor 95.48 & 92.26 & & 58.28 & 58.36 & \mejor 34.98 & 78.29\\
    
    \midrule
    \multirow{6}{*}{\texttt{KMNIST}}
    & \multicolumn{1}{l}{\texttt{MNIST}}      &  \mejor 92.29 & 89.06 & 90.57 & \seg 90.87 & & \mejor 93.58 & 91.27 & 92.14 & 92.37 & & \mejor 25.88 & 30.81 & 24.02 & 26.30\\
    & \multicolumn{1}{l}{\texttt{FMNIST}}     &  71.03 & 82.97 & \mejor 85.42 & \seg 83.40 & & 73.95 & 83.53 & \mejor 85.10 & 83.90 & & 77.10 & 45.63 & \mejor 36.01 & 50.20\\
    & \multicolumn{1}{l}{\texttt{Letters}}    &  90.43 & 89.27 & \mejor 91.29 & \seg 91.06 & & 86.09 & 85.95 & \mejor 88.27 & 87.83 & & 32.47 & 30.16 & \mejor 22.04 & 25.97\\
    & \multicolumn{1}{l}{\texttt{notMNIST}}    &  85.03 & \seg 88.69 & \mejor 89.54 & 87.22 & & \mejor 86.21 & 89.45 & 89.17 & 86.48 & & 49.23 & 29.43 & \mejor 24.96 & 34.87\\
    & \multicolumn{1}{l}{\texttt{omniglot}}   &  \mejor 95.97 & 89.14 & 91.74 & \seg 92.57 & & \mejor 96.76 & 91.01 & 93.14 & 93.74 & & \mejor 10.36 & 29.76 & 20.47 & 19.89\\
    & \multicolumn{1}{l}{\texttt{CIFAR10-BW}} &  62.77 & \seg 91.84 & \mejor 94.52 & 91.06 & & 64.71 & 93.95 & \mejor 95.99 & 93.63 & & 90.14 & 21.76 & \mejor 10.03 & 27.21\\
    \midrule
    \multirow{6}{*}{\texttt{Letters}}
    & \multicolumn{1}{l}{\texttt{MNIST}}      &  78.67 & 79.54 & \mejor 82.12 & \seg 81.83 & & 87.30 & 89.04 & \mejor 90.19 & 89.90 & & 70.08 & 77.43 & \mejor 68.02 & 70.71\\
    & \multicolumn{1}{l}{\texttt{FMNIST}}     &  87.24 & \seg 88.49 & \mejor 88.49 & 77.83 & & 94.09 & 95.03 & \mejor 95.03 & 88.41 & & 69.51 & 69.11 & \mejor 69.11 & 71.64\\
    & \multicolumn{1}{l}{\texttt{KMNIST}}     &  \mejor 92.75 & 85.46 & \seg 88.60 & 85.29 & & \mejor 96.52 & 92.52 & 93.56 & 91.91 & & \mejor 40.10 & 66.12 &  52.87 & 64.88\\
    & \multicolumn{1}{l}{\texttt{notMNIST}}    &  \mejor 87.19 & 73.97 & \seg 75.16 & 71.29 & & \mejor 93.04 & 84.58 & 85.92 & 84.25 & & \mejor 54.32 & 80.49 &  78.49 & 85.73\\
    & \multicolumn{1}{l}{\texttt{omniglot}}   &  \mejor 94.29 & 86.91 & \seg 92.90 & 92.92 & & \mejor 98.46 & 93.47 & 96.29 & 96.31 & & \mejor 30.98 & 61.63 &  34.97 & 34.46\\
    & \multicolumn{1}{l}{\texttt{CIFAR10-BW}} &  76.07 & \seg 89.94 & \mejor 89.94 & 70.92 & & \mejor 96.53 & 95.67 & 95.67 & 87.20 & & 94.32 & 66.01 & \mejor 66.01 & 97.53\\
    
    \bottomrule
    \end{tabular}}
 \end{table}
%JAVI - 22/09/2022

We now shift the focus of the discussion on the results for the CNN-SNN architecture shown in Table \ref{tb:CONVresults}. In this second experiment, our method outperforms the rest of competing OoD detectors in a wider number of cases. Exceptions are \texttt{Letters} and \texttt{KMNIST} datasets, especially the latter where the SCP-based detector only performs best in 2 out of 6 OoD datasets. Furthermore, performance gaps in those cases where it dominates the rest of the benchmark are small. As in the previously discussed results for the FC-SNN architecture, the proposed SCP-based method is unable to properly detect OoD samples in one of the OoD datasets (especifically, \texttt{CIFAR10-BW}). It is also worth noting that, although our proposal attains superior results in less cases than in the FC-SNN architectures, the differences to the winning OoD technique are not that large.

The smaller performance gaps observed in this second set of experiments call for a further examination of their statistical significance. To this end, we resort to two different methodologies widely used by the community for the purpose, and apply them to the AUROC scores reported in Tables \ref{tb:FCresults} and \ref{tb:CONVresults}: 
\begin{itemize}[leftmargin=*]
    \item A critical distance diagram \cite{demvsar2006statistical}, which employs a post-hoc Nemenyi test to compute a critical distance value; if the average rankings of two methods as per the paired score values achieved across different cases are separated by more than the critical distance value, they can be considered to perform significantly different to each other. Otherwise, if their average ranks differ in less than the critical distance, they are declared to be statistically equivalent.
    \item A Bayesian posterior analysis of the differences between the paired performance scores achieved by the techniques under comparison. The work in \cite{benavoli2017time} exposed that standard hypothesis testing was not suitable for assessing the significance of performance differences for a number of issues. Alternatively, they proposed to perform a Bayesian sign test of the performance scores of two different techniques, yielding an estimation of the probability that one technique outperforms the other using the scores obtained by each of them over all cases. The probability resulting from this analysis can be carried out via Monte Carlo sampling and depicted in a system of barycentric coordinates, in which three regions are distinguished: one where the first OoD technique outperforms the second, a second one where the second outperforms the first, and a third region reflecting \emph{practical equivalence}. This latter region delimits the probability that techniques can be concluded to perform equivalently from the statistical point of view, and depends on a parameter (\emph{rope}) that establishes the minimum difference between the scores of the OoD techniques for considering their performance as significantly different. The value of rope depends on the task being solved: in our case, \textit{rope} is set equal to $\text{AUROC}=1\%$.
\end{itemize}
\begin{figure}[hbt]
\centering
\begin{subfigure}{0.48\textwidth}
    \includegraphics[width=\textwidth]{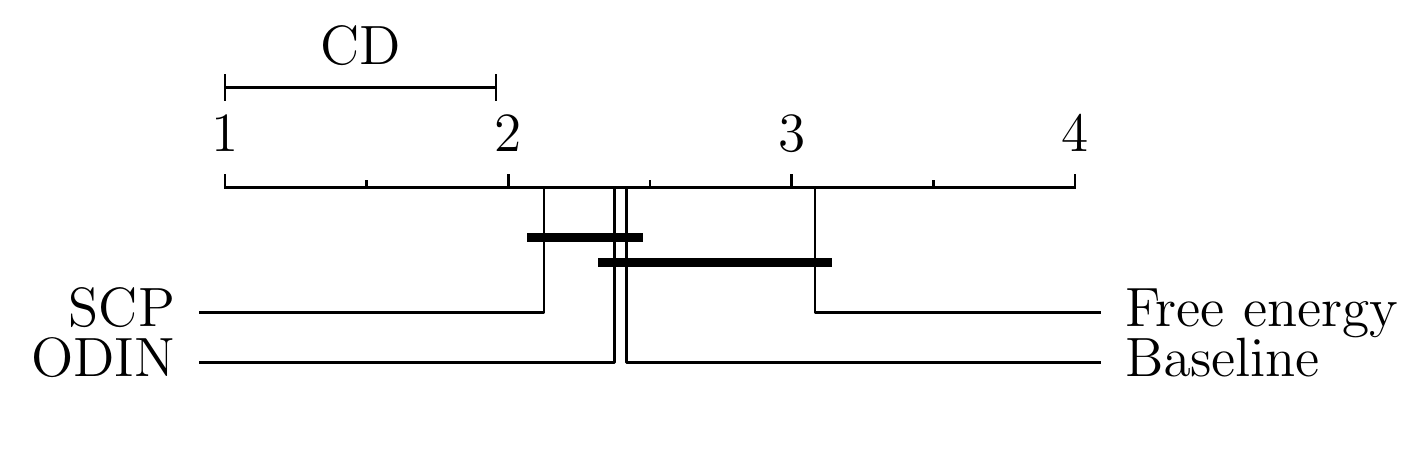}
    \caption{FC-SNN}
    \label{fig:CD_FC}
\end{subfigure}
\hfill
\begin{subfigure}{0.48\textwidth}
    \includegraphics[width=\textwidth]{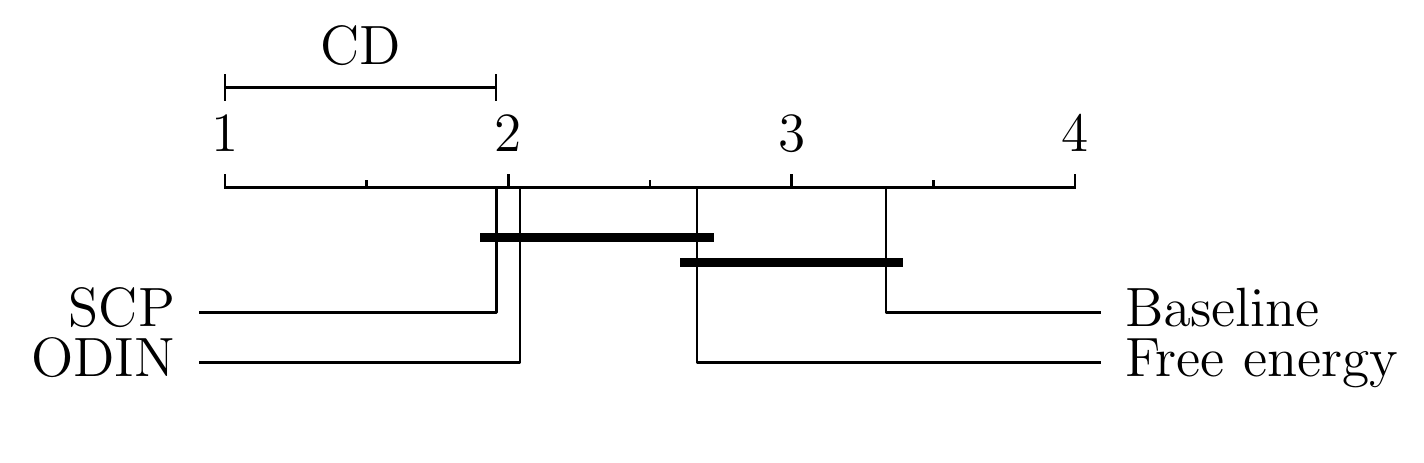}
    \caption{CNN-SNN}
    \label{fig:CD_Conv}
\end{subfigure}
\caption{Critical distance diagrams computed over the AUROC results across the cases considered in the experiments for (a) FC- and (b) CNN-SNN architectures. The value of the critical distance is shown on top of every diagram. Horizontal lines connecting the average rank markers of several detectors indicate that their average rankings cannot be claimed to be different to each other with statistical significance.}
\label{fig:CDgraphs}
\end{figure}

Figure \ref{fig:CDgraphs} depicts the result of the first statistical significance study over the results obtained for the (a) FC- and (b) CNN-SNN architectures. It is straightforward to note that for both SNN architectures, our method ranks first on average, yet its dominance over the rest of methods cannot be concluded to be statistically significant. Therefore, it is necessary to perform a closer study between our SCP-based detector and the second best one, namely, ODIN. 

\begin{figure}[h]
\centering
\begin{subfigure}{0.48\textwidth}
    \includegraphics[width=\textwidth]{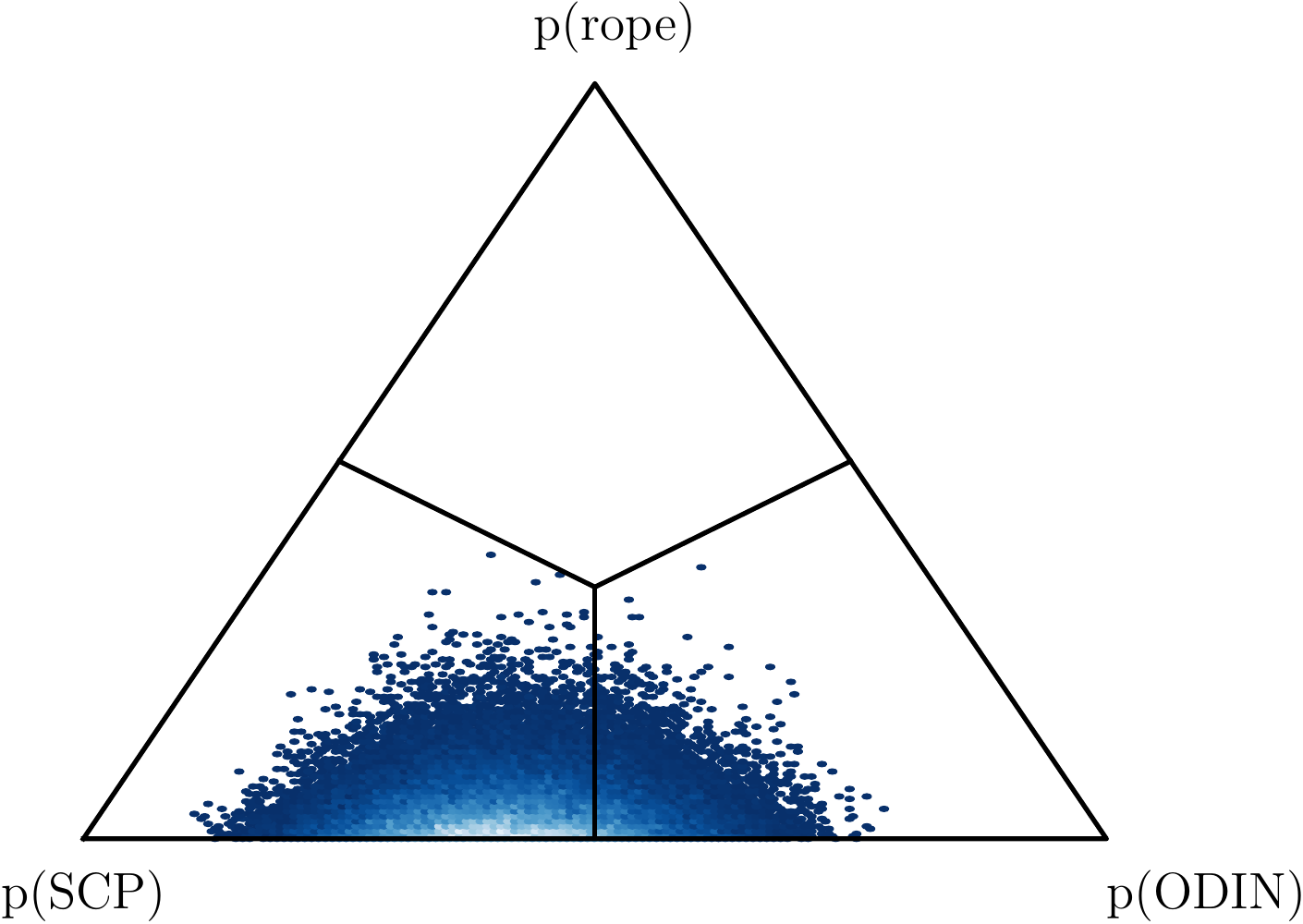}
    \caption{FC-SNN}
    \label{fig:B_FC_ODIN}
\end{subfigure}
\hfill
\begin{subfigure}{0.48\textwidth}
    \includegraphics[width=\textwidth]{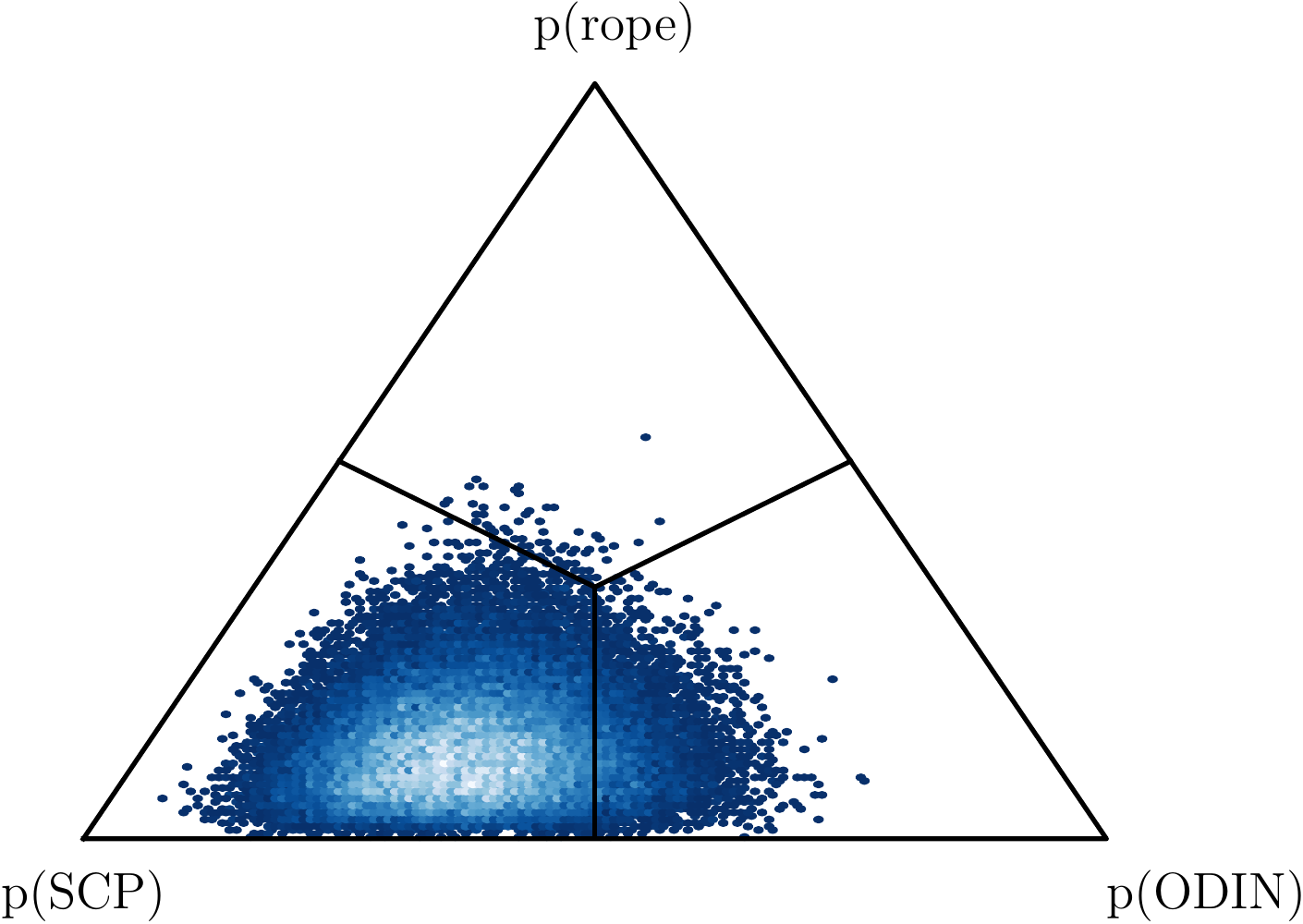}
    \caption{CNN-SNN}
    \label{fig:B_ConvODIN}
\end{subfigure}
\caption{Posterior probability resulting from the Bayesian signed tests computed over the paired AUROC scores attained by ODIN and the proposed SCP-based detector with (a) the FC-SNN architecture; (b) the CNN-SNN architecture.}
\label{fig:Bayesian}
\end{figure}

This closer examination is done by performing a Bayesian analysis of the differences following the methodology described above using the AUROC scores of ODIN and the proposed SCP-based detector. Figures \ref{fig:Bayesian}.a and \ref{fig:Bayesian}.b show the results of this second significance study. When it comes to the scores obtained for the FC-SNN architecture (Figure \ref{fig:Bayesian}.a), almost all Monte Carlo samples drawn from the adjusted Bayesian posterior probability lie in between the regions where ODIN or the SCP-based detector achieve better results. This means that none of the methods perform equivalently, and one always outperforms the other with statistical significance. This goes in line with the observation made in the discussions on Table \ref{tb:FCresults}, where it was concluded that either a logit-based approach or the proposed SCP-based detector performs best in the majority of cases, being very few the scenarios where both perform similarly. The Bayesian analysis reported in Figure \ref{fig:Bayesian}.a informs that the probability that our method performs better than ODIN is approximately $80\%$, whereas ODIN is expected to perform better for the remaining percentage.

We follow this second significance analysis by exploring the output of the Bayesian analysis over the results obtained for the CNN-SNN architecture. Figure \ref{fig:Bayesian}.b clearly reveals that either our method performs better (approximately 90\%) or ODIN does (10\%), whereas the probability that both perform equivalently (\emph{rope} region) is almost negligible.

\subsection{RQ2: Does the local relevance attribution method yield informative explanations about the features that make a given sample be detected as OoD? Which are the limitations of this explainability technique?}
\label{ssec:OoDattribution}
 
At this point we recall that the local explanation technique for the proposed SCP-based detector aims to generate a relevance attribution vector $\mathbf{h}(\mathbf{x})$ for a query sample $\mathbf{x}$ (i.e. a heatmap when dealing with image classification). The value of its components $\{h_d(\mathbf{x})\}_{d=1}^D\}$ indicates the relative relevance of regions of the input space that have influenced most the detector when detecting that the query sample is OoD. The heatmap must also deliver information about an \emph{OoD score} of the sample, such that the intensity of the heatmap should be higher as the OoD score of the sample increases, and vice-versa.

Based on the above intuition, the qualitative evaluation of the attribution method made to reply RQ2 proceeds as follows: first, attribution generated for several test images with known artifacts are shown and analyzed to verify that the produced heatmaps spot those artifacts. Next, limitations of the attribution method noted during the experimentation are exposed, using examples and results that clearly illustrate these constraints. 

We concentrate the discussion on FC-SNN and CNN-SNN architectures trained to classify \texttt{MNIST} images (in-distribution dataset), using query instances from \texttt{MNIST-Square} and \texttt{MNIST-C} as out-of-distribution datasets. The detection performance of the SCP-based OoD detector in this setup is reported in Table~\ref{tab:accuracy_att}, indicating values of the three detection scores used so far. We note that in this case the value of FPR80 is reported, i.e., the FPR value obtained when the target TPR is set to 80\%. In other words, the detection threshold is set such that 80\% of the positive (in-distribution) samples are correctly detected as such. It is important to note that all the other three logit-based detectors considered in the benchmark obtain worst detection scores in this setup. 
\begin{table}[!htbp]
\centering
\renewcommand\arraystretch{1.1}
\resizebox{0.9\textwidth}{!}{\begin{tabular}{cccccc}
    \toprule
    \makecell{ID\\dataset}& \makecell{OoD\\dataset} & Model & AUROC $\uparrow$ & AUPR $\uparrow$ & FPR80 $\downarrow$ \\
    \midrule
    \multirow{4}{*}{\texttt{MNIST}} & \multirow{2}{*}{\texttt{MNIST-Square}} & FC-SNN & 94.29 & 95.40 & 05.53 \\
    & & CNN-SNN & 76.34 & 75.32 & 44.60 \\
    \cmidrule{2-6}
    & \texttt{MNIST-C} & FC-SNN & 80.80 & 81.32 & 37.31 \\
    & \texttt{zigzag} & CNN-SNN & 75.77 & 74.04 & 43.95 \\
    \bottomrule
\end{tabular}}
\caption{Scores for the SCP-based detector in the experiments devised to address RQ2.}
\label{tab:accuracy_att}
\end{table}

\begin{figure}[htb]
\centering
\begin{subfigure}{0.48\textwidth}
    \includegraphics[width=\textwidth]{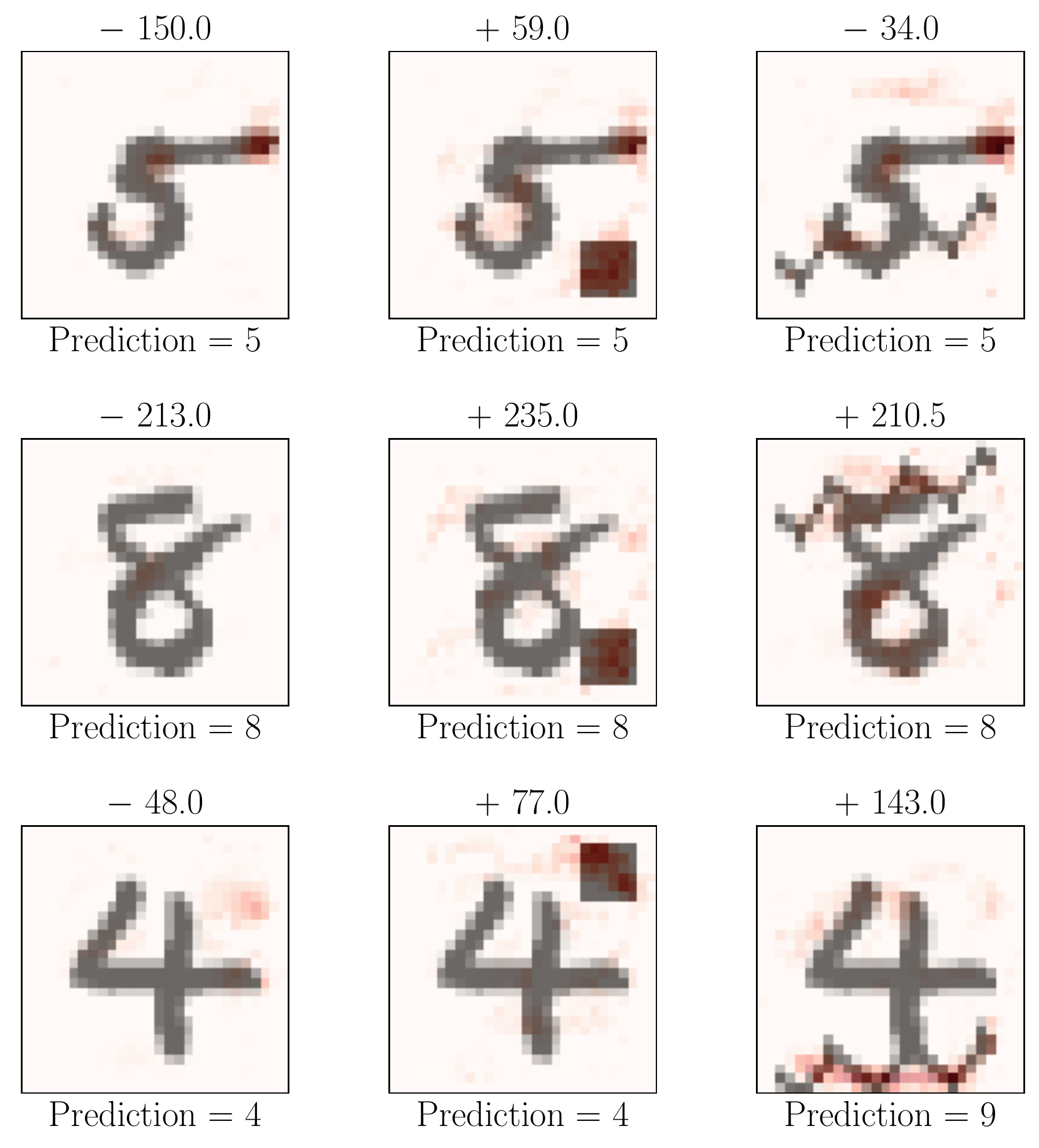}
    \caption{FC-SNN}
    \label{fig:att_FC}
\end{subfigure}
\hfill
\begin{subfigure}{0.48\textwidth}
    \includegraphics[width=\textwidth]{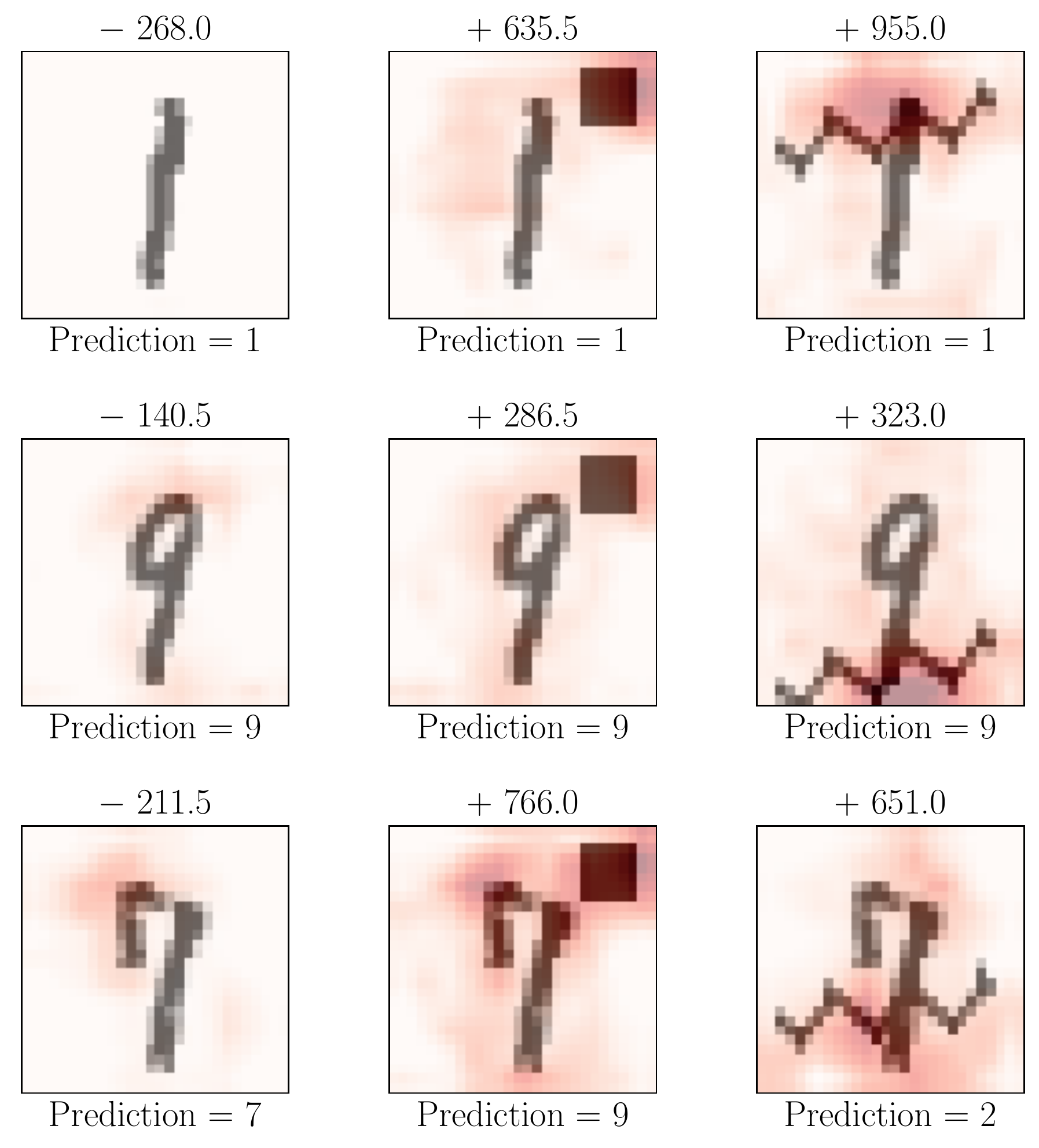}
    \caption{CNN-SNN}
    \label{fig:att_conv}
\end{subfigure}
\caption{Heatmaps showing the relevance attributed to different digits (clean and corrupted versions with square and zigzag artifacts) for (a) a FC-SNN architecture; (b) a CNN-SNN architecture. Each heatmap is superimposed on its input, and for every image its predicted class is reported, together with the difference between the its OoD score and the class-conditional threshold $\lambda^{\hat{y}}$.}
\label{fig:att_gOoD}
\end{figure}

After training both SNN architectures over the \texttt{MNIST} dataset, several in-distribution and out-of-distribution samples are tested and processed through the SCP-based detector and the local relevance attribution technique. Figure \ref{fig:att_gOoD} depicts several of such query images. The text at the bottom of each image represents the predicted class for the image, whereas the number above the image corresponds to the difference between the OoD score for the given sample -- given by the value returned by $f_{OoD}(\mathbf{x},\hat{y})$ as per Expression \eqref{eq:scoreOoD} -- and the threshold $\lambda^{\hat{y}}$ for a TPR of 80\%. Hence, a positive value above every image means that the image is detected as OoD. Conversely, the more negative this value is, the more in-distribution the sample can be thought to be, as it is \emph{farther} from being detected as OoD based on the criterion in Expression \eqref{eq:OoDdecision}.

The discussion starts with the images shown for the FC-SNN architecture (Figure \ref{fig:att_gOoD}.a). The first example, a digit corresponding to class 5, the rightmost tip of the digit is highlighted as relevant even if clearly belonging to the in-distribution. Indeed, despite its largely negative score (therefore being confidently detected as ID), usually the 5 digits in the \texttt{MNIST} dataset do not reach the edges of the image. In this case, the 5 sample depicted in the figure has an untypically long tip. In the square and zigzag case this observation is further buttressed: not only the same digit with square and zigzag artifacts preserve a highlighted right tip, but relevance is also attributed by our approach to the parts of the image affected by the artifacts. The 5 digit with the zigzag pattern is not be detected as OoD, but its OoD score is close to 0, hence the image is more likely to be an OoD than the normal case. Consequently, the produced heatmap is less intense. The 8 digit comprising square and zigzag artifacts would be detected as OoD, an both artifacts are partly remarked in the heatmaps. Some parts of the clean 8 digit image are also marked in the heatmap: this can be explained because the closest cluster for the samples with the corruption changes w.r.t. the clean digit from where they are produced, which may be composed by digits with different compositional characteristics. In the last example, the 4 digit, the zigzag case is predicted to be a digit 9 rather than a digit 4, and thus highlights several regions where the shape of an archetypical 9 could be. 

Digits shown for the CNN-SNN architecture exhibit in general noisier and smoother heatmaps, as relevance attribution vectors at the input of the architecture are scaled up via interpolation to match the dimensions of the original input image (in this case, from $11 \times 11$ to $28 \times 28$ pixels. In all examples included in Figure \ref{fig:att_gOoD}.b, the artifacts induced in the clean 1, 9 and 7 digits are highlighted in the heatmap. In the case of the digit 7, the noisier heatmaps are due to the fact that the predicted classes are 9 (square) and 2 (zigzag) rather than a 7. Hence, the parts of the image spotted by the heatmaps include both the artifacts and the parts of the original digit that make the spike count patterns deviate from those of the closest centroid of its wrongly predicted class.

In summary, we conclude that the proposed relevance attribution technique can generate heatmaps that provide meaningful information about the features of the input that make the SCP detector identify it as an out-of-distribution sample. However, during our experiments we noted several limitations of the approach that spur future research lines later described in the concluding part of the manuscript. Such limitations are: 
\begin{itemize}[leftmargin=*]

\item \emph{The depth of the network degrades the quality of attribution heatmaps}: one effect observed when applying the proposed relevance attribution technique is that heatmaps become noisier as the depth of the fully connected part of the SNN increases, i.e., the number of fully connected layers $L$. This effect can be observed in Figure \ref{fig:att_depth}, where the same inputs as in Figure \ref{fig:att_gOoD} are used over FC-SNN and CNN-SNN architectures with $L=2$ fully-connected layers, instead of the single-layered architecture considered in previous experiments.
\begin{figure}[h!]
\centering
\vspace{-3mm}
\begin{subfigure}{0.48\textwidth}
    \includegraphics[width=\textwidth]{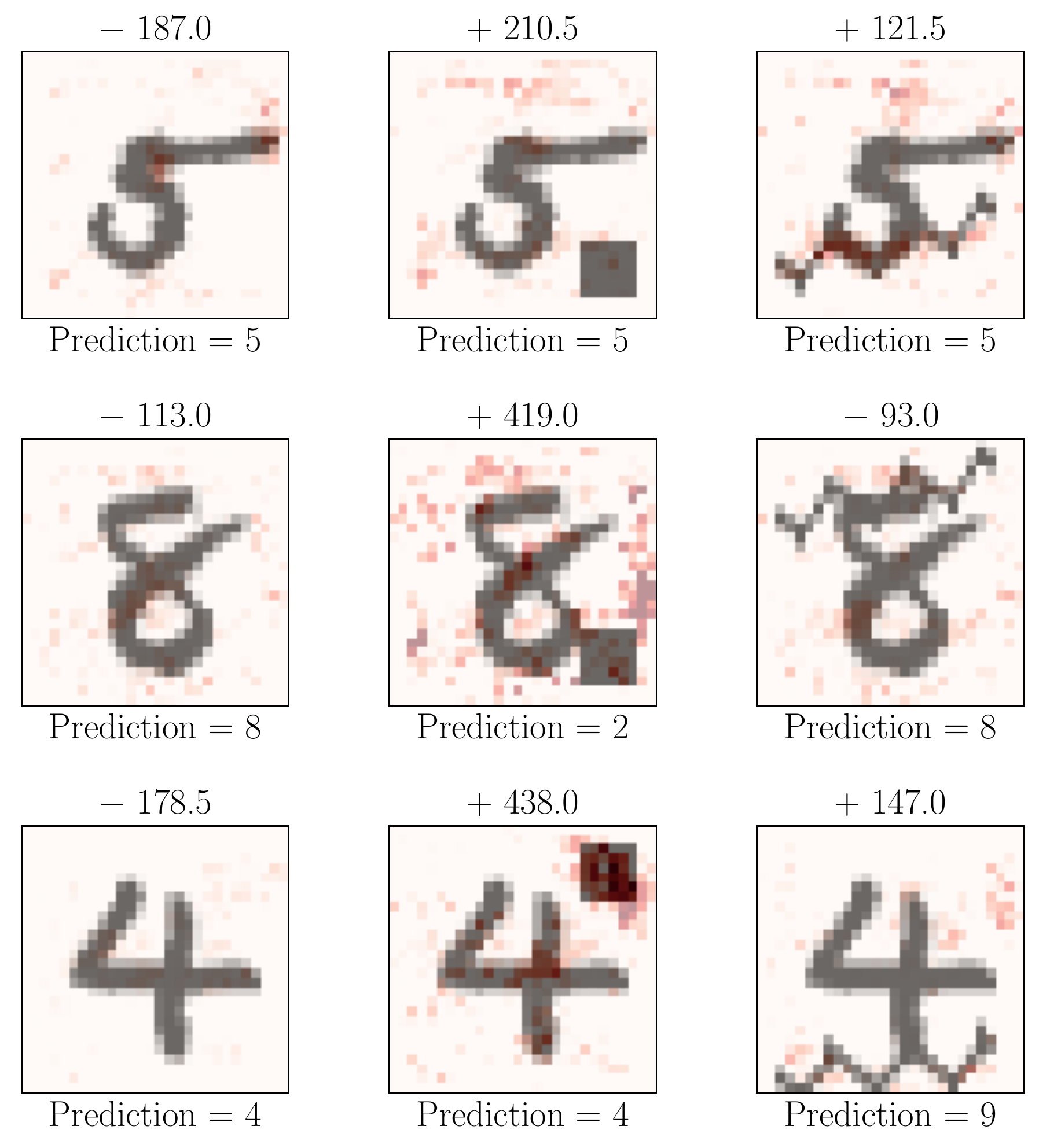}
    \caption{FC-SNN}
    \label{fig:att_FC_depth}
\end{subfigure}
\hfill
\begin{subfigure}{0.48\textwidth}
    \includegraphics[width=\textwidth]{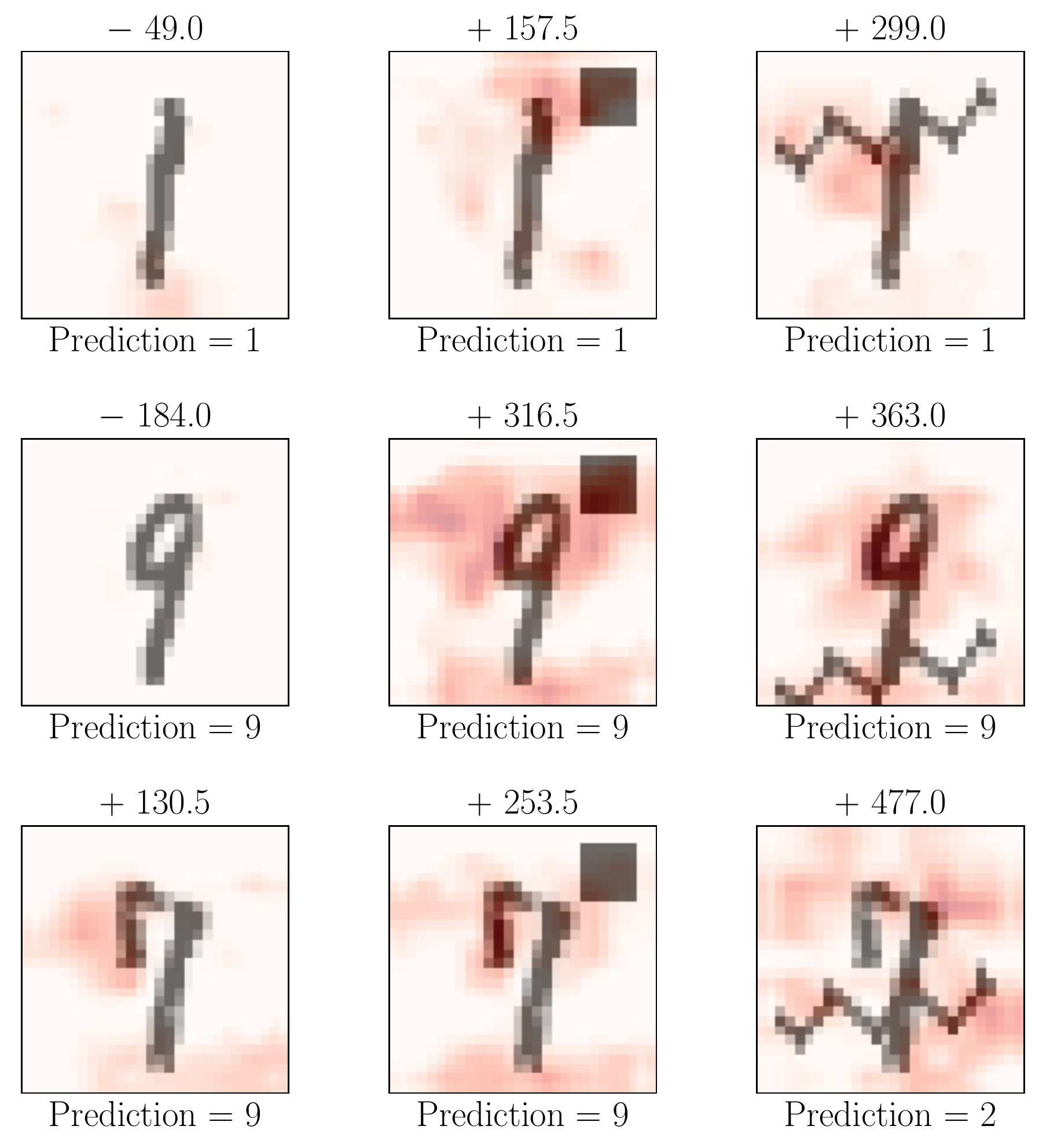}
    \caption{CNN-SNN}
    \label{fig:att_conv_depth}
\end{subfigure}
\caption{Heatmaps showing the relevance attributed to different digits (clean and corrupted versions with square and zigzag artifacts) for (a) a FC-SNN architecture; (b) a CNN-SNN architecture. In both cases the number of fully connected SNN layers is $L=2$. The layout and interpretation of the plots are identical to those in Figure \ref{fig:att_gOoD}.}
\label{fig:att_depth}
\end{figure}

As noted in these plots, in general the attribution technique elicits less sharpely defined heatmaps, evincing its lower capability to emphasize important features in the input image towards the OoD detection. Nonetheless, for some samples their predicted class has changed. Specifically, the digit 8 with the square artifact in the FC-SNN model, and the digit 7 in the CNN-SNN architecture. Moreover, the lower definition and quality of the produced attributions is not reflected on the detection scores, as can be verified in Tables \ref{tb:FCresults_depth} and \ref{tb:CONVresults_depth} and in Figure \ref{fig:Bayesian_depth} of \ref{app:depth}. While in the case of the CNN-SNN architecture a small performance loss is noted, when it comes to the FC-SNN architecture the developed SCP-based detector yields even better detection scores when compared to ODIN.

\item \emph{The attribution depends on features extracted from the input by the network}: Our detector stringently depends on the features extracted from the input image and that are input to the fully-connected part of the SNN. That is, when it is preceded by a feature extractor like a convolutional layer, our detection capabilities depend on the ability of this layer to learn feature maps that capture characteristics that push the SCP-detector to declare that the image is out-of-distribution. Unfortunately, satisfying this necessary condition is not sufficient to guarantee that OoD samples will be detected as such. A false negative could lead to a heatmap that does not highlight characteristics that expose the OoD nature of the misdetected input sample.

On the one hand, as feature maps go through a fully connected network before reaching the layer where the SCP-based detection is performed (the layer prior to softmax), the effect on the latter could be compensated for by the learned weights, as neurons are densely connected and every neuron affects all the neurons of the subsequent layers. On the other hand, due to the way the OoD score is computed, a high intra-cluster variance could eventually mask an OoD sample. As described in Algorithm \ref{alg:SCP}, our method assigns an OoD score computed as the $L_1$ distance to the closest centroid among the spike count clusters of the predicted class. Hence, if the spike count pattern of the input changes, the closest cluster to the input may vary accordingly, specially if the intra-cluster variance is high. This is not a bad effect overall, as it allows in-distribution samples to be more accurately represented by the centroids. But what could also occur is that the reconstruction of one of the centroids (mentioned in Subsection \ref{ssec:Attribution}) of the predicted class coincidentally matches the reconstruction of the input being tested, masking the OoD characteristics that should be highlighted as relevant.
\end{itemize}

Notwithstanding the above limitations, results reported in this section support that i) the proposed SCP-based OoD detector performs very competitively with respect to other model-agnostic OoD detection approaches from the literature, with significance as per a Bayesian analysis of the performance differences; and ii) that the relevance attribution technique effectively highlights artificially induced modifications to in-distribution data that a priori should contribute to their identification as OoD by the SCP-based detector. 

\section{Conclusions and future work}
\label{sec:Conclusions}

This work has elaborated on the out-of-distribution detection problem when dealing with spiking neural network architectures. Focusing on networks with rate-based encoding schemes and leaky integrate-and-fire neurons, the main idea of the proposed OoD detector is to leverage the spike count patterns produced at the last intermediate layer of the neural model to create archetypes or representatives of those patterns for every class in the dataset at hand. By comparing these archetypes with spike count patterns stimulated by new query samples, the proposed detector can effectively discriminate whether they are in- or out-distributions samples. Comparisons are made by using a distance metric, whereas archetypes are represented by the centroids of a cluster space computed over the spike count patterns of training samples by a clustering approach. We further inform the output of the OoD detector with explanations issued by a local relevance attribution method, providing a measure of the features of any given input sample that drive most the output of the developed detector towards declaring the input as out-of-distribution. 

Extensive experiments done for image classification tasks over fully-connected and convolutional-based SNN architectures have verified that the proposed detector achieves a competitive detection performance over other model-agnostic post-hoc OoD detection techniques, which have been tailored to work with spiking-based neural models. Furthermore, explanations issued by the proposed local relevance attribution method excel at identifying artificially inserted perturbations on in-distribution samples, which are highlighted as the parts of the image that are influential for their detection as OoD. Limitations of this technique have been exposed and discussed, showing i) the degradation of the quality of explanations with the increasing depth of the fully connected part of the network; ii) the dependence of the attributed relevance on the input features captured by the input of the SNN model; and iii) the intra-cluster variance of clusters computed from the spike counts of the input data.

Several research directions are envisioned for the near future arising from the promising results presented in this work. To begin with, we plan to extend the applicability of the SCP detector and the principles on which its design relies to other types of encoding methods suitable for spiking neural architectures. Efforts will be also invested towards mitigating the impact of the aforementioned limitations on the produced heatmaps of the relevance attribution technique. Other forms of explanatory information about the detection of OoD samples will be explored, such as the use of generative spiking models for counterfactual studies of the audited SNN model. Finally, OoD detection in temporal or spatio-temporal streaming data, such as
those captured by dynamic vision sensors \cite{paulun2018retinotopic}, brain EEG and fMRI data \cite{KASABOV2019341}, seismic data \cite{kasabov2019time} or financial data \cite{AbouHassan:22} is another direction that we plan to follow in future studies.

\section*{Acknowledgments}

A. Martinez Seras receives funding support from the Basque Government through its BIKAINTEK PhD support program. J. Del Ser acknowledges funding support from the same institution through the Consolidated Research Group MATHMODE (IT1456-22) and the ELKARTEK program (EGIA, grant no. KK-2022/00119). P. García Bringas also thanks the funding support from the Basque Government through the Consolidated Research Group D4K - Deusto for Knowledge  (IT1528-22) and the ELKARTEK funding grants REMEDY (ref. KK-2021/00091) and SIIRSE (ref. KK-2022/00029).

\newpage
\appendix

\section{OoD detection results for SNN with increased depth} \label{app:depth}

\begin{table}[!hp]
\caption{OoD detection scores corresponding to the FC-SNN architecture with $L=2$ fully connected layers.}
\label{tb:FCresults_depth}
\centering
\renewcommand\arraystretch{0.85}
\resizebox{\textwidth}{!}{\begin{tabular}{cccccccccccccccc}
    & & \multicolumn{4}{c}{AUROC $\uparrow$} & & \multicolumn{4}{c}{AUPR $\uparrow$} & & \multicolumn{4}{c}{FPR95 $\downarrow$} \\
    \cmidrule{3-6} \cmidrule{8-11} \cmidrule{13-16}
    \makecell{ID\\dataset}& \makecell{OoD\\dataset} & \makecell{SCP\\(proposed)} & Baseline & ODIN & Energy & &   \makecell{SCP\\(proposed)} & Baseline & ODIN & Energy & &  \makecell{SCP\\(proposed)} & Baseline & ODIN & Energy  \\
    \midrule
    \multirow{6}{*}{\texttt{MNIST}}
    & \multicolumn{1}{l}{\texttt{FMNIST}}     &  \mejor 87.92 & 85.38 & \seg 86.97 & 86.15 & & \mejor 89.42 & 83.45 & 86.24 & 86.89 & & \mejor 62.19 & 59.78 &  58.59 & 67.40\\
    & \multicolumn{1}{l}{\texttt{KMNIST}}     &  \mejor 92.35 & 91.17 & \seg 91.93 & 87.64 & & \mejor 92.85 & 91.69 & 92.04 & 91.11 & & \mejor 38.58 & 50.62 &  43.88 & 48.21\\
    & \multicolumn{1}{l}{\texttt{Letters}}    &  \mejor 87.64 & 84.09 & \seg 84.44 & 83.34 & & \mejor 77.61 & 66.66 & 64.83 & 63.83 & & \mejor 51.49 & 61.48 &  56.61 & 63.25\\
    & \multicolumn{1}{l}{\texttt{notMNIST}}    &  \mejor 97.56 & 82.02 & \seg 82.02 & 77.00 & & \mejor 97.78 & 80.40 & 80.40 & 73.95 & & \mejor 11.73 & 69.20 &  69.20 & 79.53\\
    & \multicolumn{1}{l}{\texttt{omniglot}}   &  84.42 & 95.14 & \seg 97.22 & \mejor 97.58 & & 84.72 & 95.75 & 97.30 & \mejor 97.63 & & 64.93 & 32.69 &  15.10 & \mejor 13.05\\
    & \multicolumn{1}{l}{\texttt{CIFAR10-BW}} &  \mejor 92.49 & 85.31 & \seg 86.11 & 84.59 & & \mejor 93.71 & 86.77 & 87.37 & 86.18 & & \mejor 42.27 & 71.26 &  68.98 & 75.37\\
    \midrule
    \multirow{6}{*}{\texttt{FMNIST}}
    & \multicolumn{1}{l}{\texttt{MNIST}}      &  \seg 94.23 & 83.23 & 93.62 & \mejor 95.65 & & 94.17 & 85.59 & 94.16 & \mejor 95.81 & & 24.92 & 65.76 &  33.74 & \mejor 20.86\\
    & \multicolumn{1}{l}{\texttt{KMNIST}}     &  \mejor 88.26 & 77.06 & 85.23 & \seg 86.39 & & \mejor 87.65 & 79.23 & 82.65 & 86.64 & & \mejor 47.76 & 77.98 &  63.28 & 59.90\\
    & \multicolumn{1}{l}{\texttt{Letters}}    &  \mejor 93.63 & 81.14 & 90.12 & \seg 91.53 & & \mejor 89.10 & 74.17 & 84.95 & 86.86 & & \mejor 30.71 & 71.11 &  50.21 & 44.41\\
    & \multicolumn{1}{l}{\texttt{notMNIST}}    &  \mejor 94.32 & 67.80 & 76.85 & \seg 81.60 & & \mejor 94.56 & 71.25 & 77.64 & 83.93 & & \mejor 26.87 & 85.03 &  77.54 & 90.52\\
    & \multicolumn{1}{l}{\texttt{omniglot}}   &  87.77 & 87.12 & \seg 94.87 & \mejor 96.60 & & 86.26 & 87.51 & 94.64 & \mejor 96.42 & & 46.90 & 53.31 &  24.11 & \mejor 14.74\\
    & \multicolumn{1}{l}{\texttt{CIFAR10-BW}} & \seg 78.77 & 52.32 & 68.25 & \mejor 85.47 & & 81.64 & 64.84 & 74.69 & \mejor 89.82 & & \mejor 76.82 & 96.00 & 92.18 & 94.79\\
    
    \midrule
    \multirow{6}{*}{\texttt{KMNIST}}
    & \multicolumn{1}{l}{\texttt{MNIST}}      &  77.27 & 84.17 & \mejor 85.39 & \seg 84.82 & & 78.26 & 86.42 & \mejor 87.37 & 86.99 & & 70.25 & 51.88 & \mejor 44.86 & 47.97\\
    & \multicolumn{1}{l}{\texttt{FMNIST}}     & \seg 78.13 & 84.72 & \mejor 84.72 & 76.98 & & 79.30 & 87.92 & \mejor 87.92 & 81.95 & & 70.52 & 53.21 & \mejor 53.21 & 72.84\\
    & \multicolumn{1}{l}{\texttt{Letters}}    &  83.10 & 84.02 & \mejor 85.10 & \seg 84.05 & & 73.55 & 78.36 & \mejor 79.65 & 78.67 & & 60.62 & 53.18 & \mejor 47.54 & 51.95\\
    & \multicolumn{1}{l}{\texttt{notMNIST}}    &  \mejor 95.89 & \seg 79.13 &  79.13 & 74.21 & & \mejor 96.11 & 81.36 & 81.36 & 76.96 & & \mejor 14.31 & 62.46 & 62.46 & 76.91\\
    & \multicolumn{1}{l}{\texttt{omniglot}}   &  76.72 & 92.00 & \seg 94.63 & \mejor 95.47 & & 77.42 & 93.50 & 95.71 & \mejor 96.38 & & 74.59 & 26.58 & 12.52 & \mejor 09.33\\
    & \multicolumn{1}{l}{\texttt{CIFAR10-BW}} & \mejor 92.96 & \seg 89.81 &  90.58 & 91.00 & & \mejor 93.37 & 92.01 & 92.65 & 91.03 & & \mejor 26.84 & 35.71 & 28.27 & 43.33\\
    \midrule
    \multirow{6}{*}{\texttt{Letters}}
    & \multicolumn{1}{l}{\texttt{MNIST}}      &  73.49 & 80.03 & \seg 81.95 & \mejor 82.06 & & 83.87 & 87.82 & \mejor 88.40 & 88.38 & & 81.48 & 71.45 & \mejor 61.37 & 61.60\\
    & \multicolumn{1}{l}{\texttt{FMNIST}}     &  \mejor 86.89 & 81.15 & \seg 81.15 & 65.36 & & \mejor 92.65 & 90.17 & 90.17 & 81.67 & & \mejor 51.56 & 76.88 &  76.88 & 88.14\\
    & \multicolumn{1}{l}{\texttt{KMNIST}}     &  \mejor 88.66 & 85.55 & \seg 84.06 & 81.42 & & \mejor 93.99 & 90.64 & 91.20 & 89.74 & & \mejor 50.03 & 70.53 &  67.51 & 72.93\\
    & \multicolumn{1}{l}{\texttt{notMNIST}}    &  \mejor 98.28 & 74.10 & \seg 74.21 & 66.57 & & \mejor 99.12 & 83.82 & 83.82 & 78.62 & & \mejor 08.25 & 80.33 &  80.33 & 82.07\\
    & \multicolumn{1}{l}{\texttt{omniglot}}   &  84.64 & 92.76 & \seg 96.91 & \mejor 97.13 & & 92.52 & 96.39 & 98.47 & \mejor 98.56 & & 74.26 & 37.70 &  16.79 & \mejor 14.90\\
    & \multicolumn{1}{l}{\texttt{CIFAR10-BW}} &  \mejor 98.08 & 70.88 & \seg 70.88 & 50.07 & & \mejor 99.14 & 85.39 & 85.39 & 74.36 & & \mejor 09.66 & 91.43 &  91.43 & 98.70\\
    
    \bottomrule
    \end{tabular}}
 \end{table}
\begin{table}[!htbp]
\caption{OoD detection scores corresponding to the CNN-SNN architecture with $L=2$ fully connected layers.}
\label{tb:CONVresults_depth}
\centering
\renewcommand\arraystretch{0.85}
\resizebox{\textwidth}{!}{\begin{tabular}{cccccccccccccccc}
    & & \multicolumn{4}{c}{AUROC $\uparrow$} & & \multicolumn{4}{c}{AUPR $\uparrow$} & & \multicolumn{4}{c}{FPR95 $\downarrow$} \\
    \cmidrule{3-6} \cmidrule{8-11} \cmidrule{13-16}
    \makecell{ID\\dataset}& \makecell{OoD\\dataset} & \makecell{SCP\\(proposed)} & Baseline & ODIN & Energy & &   \makecell{SCP\\(proposed)} & Baseline & ODIN & Energy & &  \makecell{SCP\\(proposed)} & Baseline & ODIN & Energy  \\
    \midrule
    \multirow{6}{*}{\texttt{MNIST}}
    & \multicolumn{1}{l}{\texttt{FMNIST}}     &  \mejor 95.53 & 88.22 & \seg 89.92 & 88.66 & & \mejor 96.09 & 88.00 & 90.32 & 91.20 & & \mejor 22.60 & 54.07 &  45.15 & 50.01\\
    & \multicolumn{1}{l}{\texttt{KMNIST}}     &  \mejor 95.57 & 90.62 & \seg 91.56 & 91.19 & & \mejor 95.94 & 90.18 & 90.72 & 90.25 & & \mejor 22.78 & 46.76 &  41.06 & 41.62\\
    & \multicolumn{1}{l}{\texttt{Letters}}    &  \mejor 88.60 & 81.65 & \seg 81.69 & 80.98 & & \mejor 74.99 & 59.88 & 58.44 & 57.42 & & \mejor 39.25 & 57.96 &  56.24 & 58.31\\
    & \multicolumn{1}{l}{\texttt{notMNIST}}    &  \mejor 96.40 & 85.75 & \seg 86.85 & 87.01 & & \mejor 96.63 & 82.94 & 84.22 & 84.42 & & \mejor 18.08 & 54.51 &  51.83 & 51.85\\
    & \multicolumn{1}{l}{\texttt{omniglot}}   &  \mejor 95.35 & 90.87 & 92.17 & \seg 92.23 & & \mejor 95.61 & 90.53 & 91.37 & 91.33 & & \mejor 23.77 & 44.74 &  37.33 & 36.10\\
    & \multicolumn{1}{l}{\texttt{CIFAR10-BW}} &  \mejor 97.60 & 94.60 & \seg 96.04 & 95.01 & & \mejor 98.09 & 95.13 & 96.15 & 92.77 & & \mejor 10.76 & 32.01 &  21.44 & 28.23\\
    \midrule
    \multirow{6}{*}{\texttt{FMNIST}}
    & \multicolumn{1}{l}{\texttt{MNIST}}      &  \mejor 93.84 & 78.07 & 90.43 & \seg 92.86 & & \mejor 93.53 & 80.92 & 91.12 & 93.21 & & \mejor 25.82 & 73.54 &  42.41 & 33.24\\
    & \multicolumn{1}{l}{\texttt{KMNIST}}     &  \mejor 93.84 & 78.80 & 88.44 & \seg 90.08 & & \mejor 94.07 & 81.15 & 88.72 & 90.87 & & \mejor 30.97 & 71.66 &  47.79 & 38.00\\
    & \multicolumn{1}{l}{\texttt{Letters}}    &  \mejor 91.21 & 76.17 & 87.40 & \seg 89.61 & & \mejor 83.60 & 68.71 & 80.78 & 83.48 & & \mejor 38.28 & 77.06 &  52.06 & 45.20\\
    & \multicolumn{1}{l}{\texttt{notMNIST}}    &  \mejor 93.85 & 77.21 & \seg 86.17 & 86.14 & & \mejor 93.24 & 77.13 & 83.65 & 83.31 & & \mejor 26.12 & 75.88 &  53.79 & 54.21\\
    & \multicolumn{1}{l}{\texttt{omniglot}}   &  \mejor 92.92 & 77.54 & 88.42 & \seg 91.37 & & 90.19 & 80.90 & 89.36 & \mejor 91.88 & & \mejor 28.52 & 76.16 &  50.49 & 40.47\\
    & \multicolumn{1}{l}{\texttt{CIFAR10-BW}} &  88.11 & 82.43 & \seg 88.78 & \mejor 89.79 & & 88.60 & 84.47 & 89.28 & \mejor 92.13 & & 55.63 & 67.75 & 48.77 & \mejor 45.48\\
    
    \midrule
    \multirow{6}{*}{\texttt{KMNIST}}
    & \multicolumn{1}{l}{\texttt{MNIST}}      &  89.80 & 89.09 & \mejor 90.64 & \seg 90.53 & & 91.61 & 90.58 & \mejor 91.84 & 91.74 & & 27.37 & 33.22 & \mejor 23.42 & 24.17\\
    & \multicolumn{1}{l}{\texttt{FMNIST}}     &  84.09 & 88.29 & \seg 91.36 & \mejor 91.97 & & 87.42 & 91.04 & 93.30 & \mejor 93.71 & & 52.00 & 41.47 & 24.45 & \mejor 23.51\\
    & \multicolumn{1}{l}{\texttt{Letters}}    &  90.85 & 88.25 & \mejor 91.01 & \seg 90.99 & & \mejor 89.06 & 82.75 & 84.71 & 84.67 & & \mejor 18.83 & 36.71 & 25.85 & 25.92\\
    & \multicolumn{1}{l}{\texttt{notMNIST}}    &  88.24 & 88.33 & \mejor 92.27 & \seg 89.15 & & 89.49 & 89.38 & \mejor 91.12 & 90.84 & & \mejor 21.09 & 37.27 & 29.72 & 31.13\\
    & \multicolumn{1}{l}{\texttt{omniglot}}   &  \mejor 91.90 & 83.75 & 91.63 & \seg 92.05 & & \mejor 93.51 & 92.07 & 92.78 & 93.14 & & \mejor 19.25 & 33.35 & 20.80 & 19.83\\
    & \multicolumn{1}{l}{\texttt{CIFAR10-BW}} &  86.11 & \seg 93.99 & \mejor 96.28 & 97.49 & & 89.99 & 95.21 & \mejor 97.15 & 95.25 & & 49.31 & 13.88 & \mejor 04.92 & 03.66\\
    \midrule
    \multirow{6}{*}{\texttt{Letters}}
    & \multicolumn{1}{l}{\texttt{MNIST}}      &  74.39 & 80.75 & \seg 83.65 & \mejor 84.08 & & 89.25 & 89.35 & 90.05 & \mejor 91.05 & & 75.88 & 73.81 & \mejor 61.83 & 63.69\\
    & \multicolumn{1}{l}{\texttt{FMNIST}}     &  \mejor 86.49 & 84.88 & \seg 84.88 & 74.20 & & \mejor 93.51 & 92.63 & 92.63 & 85.59 & & \mejor 66.97 & 70.32 &  70.32 & 73.83\\
    & \multicolumn{1}{l}{\texttt{KMNIST}}     &  \mejor 90.36 & 85.05 & \seg 87.38 & 84.00 & & \mejor 98.16 & 92.56 & 93.02 & 91.01 & & \mejor 47.80 & 66.44 &  55.88 & 63.06\\
    & \multicolumn{1}{l}{\texttt{notMNIST}}    &  \mejor 86.83 & 76.21 & \seg 76.55 & 72.18 & & \mejor 95.17 & 86.57 & 86.40 & 84.20 & & \mejor 52.48 & 80.37 &  77.20 & 80.71\\
    & \multicolumn{1}{l}{\texttt{omniglot}}   &  \mejor 92.08 & 87.08 & \seg 91.39 & 91.16 & & \mejor 96.03 & 93.18 & 95.19 & 95.01 & & \mejor 38.39 & 58.70 &  39.68 & 38.95\\
    & \multicolumn{1}{l}{\texttt{CIFAR10-BW}} &  \mejor 94.27 & 88.20 & \seg 88.20 & 73.09 & & \mejor 97.62 & 94.87 & 94.87 & 88.40 & & \mejor 39.12 & 72.80 &  72.80 & 96.77\\
    
    \bottomrule
    \end{tabular}}
\end{table}
\newpage
\begin{figure}[ht]
\centering
\begin{subfigure}{0.48\textwidth}
    \includegraphics[width=\textwidth]{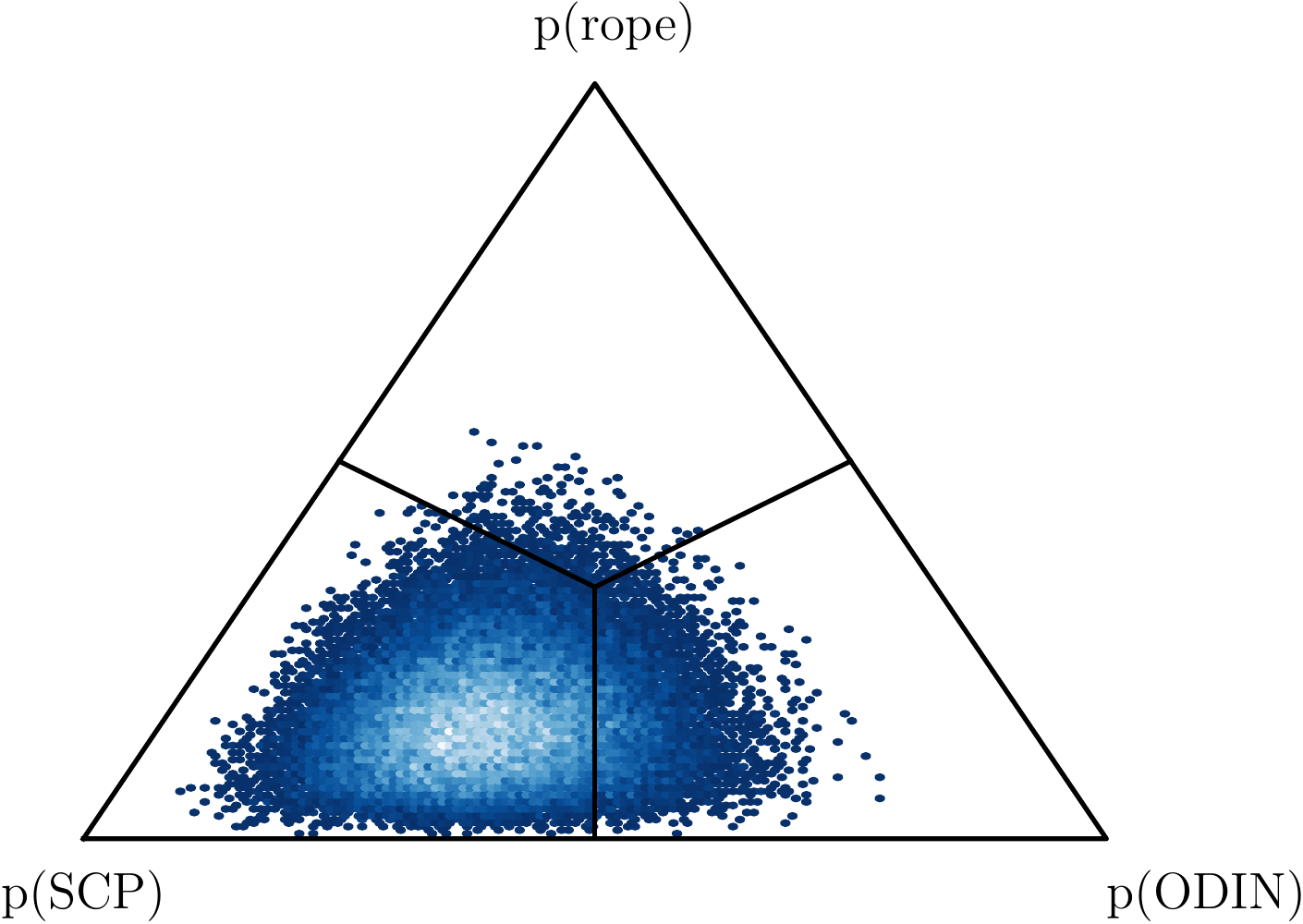}
    \caption{FC-SNN}
    \label{fig:B_FC_ODIN_depth}
\end{subfigure}
\hfill
\begin{subfigure}{0.48\textwidth}
    \includegraphics[width=\textwidth]{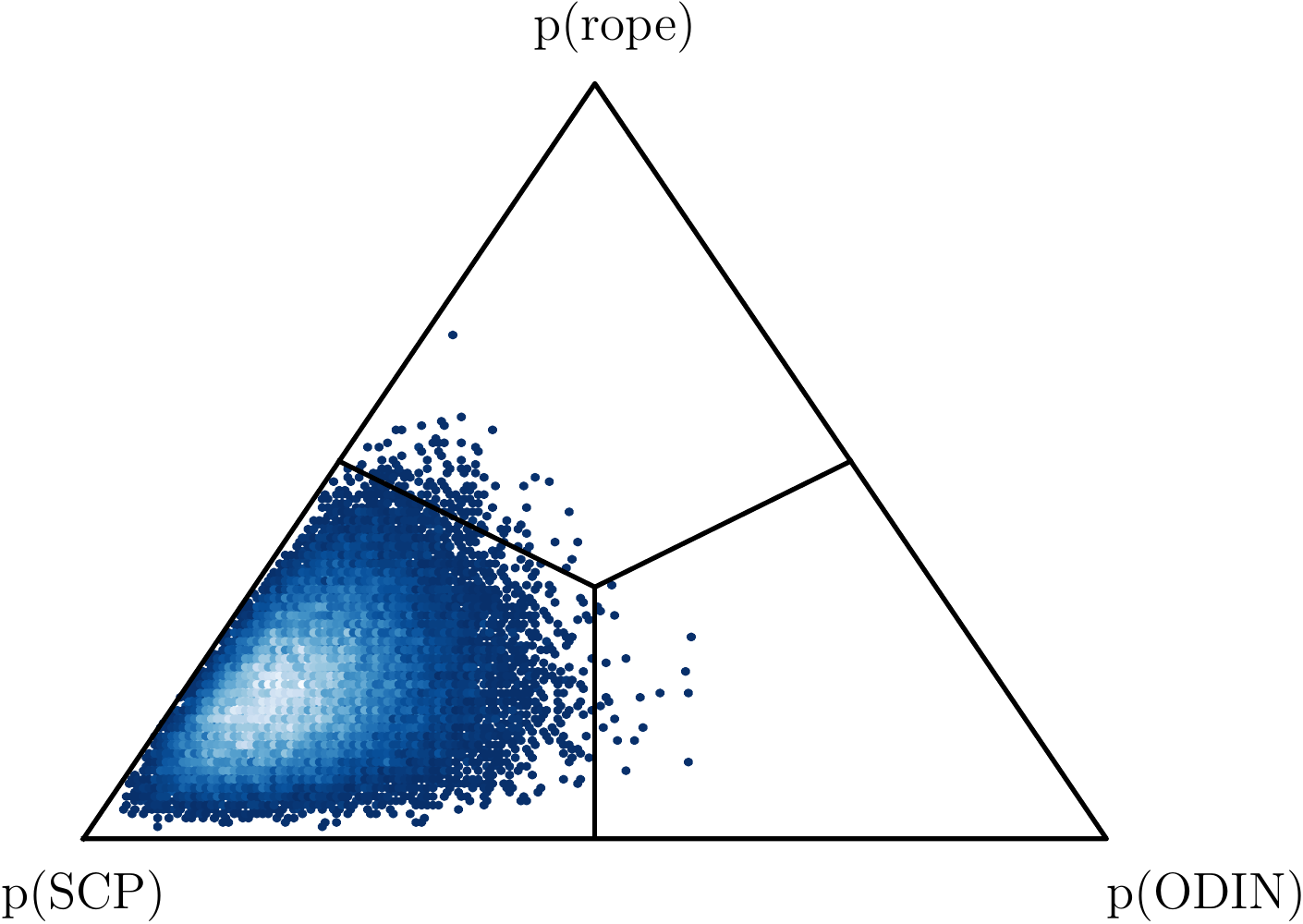}
    \caption{CNN-SNN}
    \label{fig:B_ConvODIN_depth}
\end{subfigure}
\caption{Posterior probability resulting from the Bayesian signed tests computed over the paired AUROC scores attained by ODIN and the proposed SCP-based detector with (a) the FC-SNN architecture; (b) the CNN-SNN architecture. In both cases $L=2$.}
\label{fig:Bayesian_depth}
\end{figure}

\bibliographystyle{IEEEtran}
\bibliography{bib}
\vfill

\end{document}